\pdfoutput=1

\documentclass[11pt]{article}

\usepackage[final]{acl}

\usepackage{times}
\usepackage{latexsym}
\usepackage{amsmath}

\usepackage[T1]{fontenc}

\usepackage[utf8]{inputenc}

\usepackage{microtype}

\usepackage{inconsolata}
\usepackage{graphicx}
\usepackage{bbm}
\usepackage{amsmath}
\usepackage{makecell}
\usepackage{multirow}
\usepackage{booktabs}
\usepackage{tabularx}
\usepackage{hyperref}
\usepackage{amssymb}
\usepackage{subfigure}
\usepackage{subcaption}
\usepackage{tcolorbox}
\tcbuselibrary{skins}
\usepackage{color}
\usepackage{longtable}
\usepackage{tikz,xcolor}
\usepackage{tabularx}
\usepackage{algorithm}
\usepackage{algpseudocode}
\usepackage{stmaryrd}
\usepackage{array}
\usepackage{makecell}
\newcommand{\sepsmall}{\hskip 3pt}

\definecolor{airforceblue}{rgb}{0.14, 0.31, 0.5}
\definecolor{brightmaroon}{rgb}{0.76, 0.13, 0.28}
\definecolor{mdgreen}{rgb}{0.05,0.6,0.05}
\definecolor{mdairforceblue}{rgb}{0.1, 0.3, 0.7}
\definecolor{mdblue}{rgb}{0,0,0.7}

\newcommand{\LineComment}[1]{\hfill$\triangleright$~#1}
\newcommand{\smallred}[1]{\small \textcolor{brightmaroon}{#1}}
\newcommand{\smallgreen}[1]{\small \textcolor{mdgreen}{#1}}
 \newcommand{\blue}[1]{ \textcolor{mdairforceblue}{#1}}
\newcommand{\red}[1]{\textcolor{brightmaroon}{#1}}
\newcommand{\green}[1]{ \textcolor{mdgreen}{#1}}

\title{How Do LLMs and VLMs Understand Viewpoint Rotation Without Vision?\\An Interpretability Study}

\author{
 \textbf{Zhen Yang\textsuperscript{1}},
 \textbf{Ping Jian\thanks{Corresponding author.}\textsuperscript{1}},
 \textbf{Zhongbin Guo\textsuperscript{1}},
 \textbf{Zuming Zhang\textsuperscript{1}},\\
 \textbf{Chengzhi Li\textsuperscript{1}},
 \textbf{Yonghong Deng\textsuperscript{1}},
 \textbf{Xinyue Zhang\textsuperscript{1}},
 \textbf{Wenpeng Lu\textsuperscript{2}}
\\
 \textsuperscript{1}School of Computer Science and Technology, Beijing Institute of Technology, Beijing, China \\
 \textsuperscript{2}Key Laboratory of Computing Power Network and Information Security, Ministry of\\Education,
        Shandong Computer Science Center (National Supercomputer Center in Jinan), \\
        Qilu University of Technology (Shandong Academy of Sciences), Jinan, China
\\
   \texttt{\{bityangzhen, pjian\}@bit.edu.cn} 
}

\begin{document}
\maketitle
\begin{abstract}
Over the past year, spatial intelligence has drawn increasing attention. Many prior works study it from the perspective of \emph{visual}-spatial intelligence, where models have access to visuospatial information from visual inputs. However, in the absence of visual information, whether linguistic intelligence alone is sufficient to endow models with spatial intelligence, and how models perform relevant tasks with \emph{text}-only inputs still remain unexplored. Therefore, in this paper, we focus on a fundamental and critical capability in spatial intelligence from a linguistic perspective: viewpoint rotation understanding (VRU). Specifically, LLMs and VLMs are asked to infer their final viewpoint and predict the corresponding observation in an environment given textual description of viewpoint rotation and observation over multiple steps. We find that both LLMs and VLMs perform poorly on our proposed dataset while human can easily achieve 100\% accuracy, indicating a substantial gap between current model capabilities and the requirements of spatial intelligence. To uncover the underlying mechanisms, we conduct a layer-wise probing analysis and head-wise causal intervention. Our findings reveal that although models encode viewpoint information in the hidden states, they appear to struggle to bind the viewpoint position with corresponding observation, resulting in a hallucination in final layers. Finally, we selectively fine-tune the key attention heads identified by causal intervention to improve VRU performance. Experimental results demonstrate that such selective fine-tuning achieves improved VRU performance while avoiding catastrophic forgetting of generic abilities \footnote{Our dataset and code will be released at \url{https://github.com/Young-Zhen/VRU_Interpret}}. 
\end{abstract}

\section{Introduction}
With the rapid development of large language models (LLMs) and vision-language models (VLMs), spatial intelligence has attracted increasing attention \citep{DBLP:journals/corr/abs-2504-09848}. 
Spatial intelligence generally involves perceiving and mentally manipulating spatial relationships \citep{DBLP:conf/cvpr/YangYGH0X25}, and requires myriad capabilities such as relational reasoning \citep{DBLP:conf/nips/WangMSVW0J24}, distance estimation \citep{DBLP:journals/corr/abs-2503-22976}, viewpoint transformation between egocentric and allocentric perspectives \citep{DBLP:journals/corr/abs-2505-21500, xue2025reasoningpathlatentstate}, etc. 
\begin{figure}
  \centering
  \includegraphics[width=0.48\textwidth]{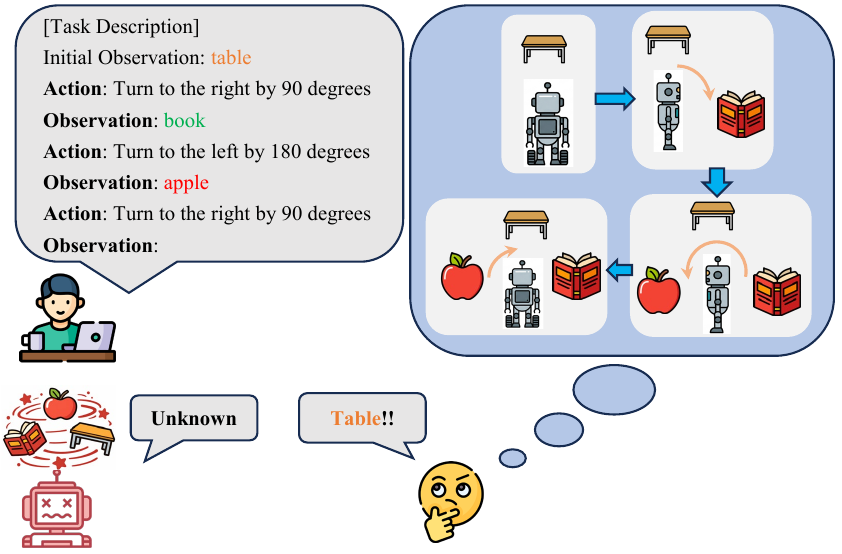}
  \caption{For textual viewpoint rotation understanding, human can easily imagine the spatial scenario and achieve 100\% accuracy, while LLMs and VLMs struggle to output the correct answer.\label{fig:illustration}}
\end{figure}
In this work, we focus on a fundamental and critical capability in spatial intelligence: \textbf{viewpoint rotation understanding} (VRU) \citep{DBLP:journals/tomccap/ZhangWZYLY25}. Specifically, we investigate whether models can accurately infer the final position and the corresponding observation in an environment, after undergoing a sequence of viewpoint rotations and receiving observations. 

Many prior studies \citep{DBLP:conf/cvpr/YangYGH0X25, DBLP:journals/corr/abs-2505-21500, DBLP:conf/acl/ZhangNMWZKLW25, DBLP:journals/corr/abs-2412-07825} have taken efforts on benchmarking and improving the spatial intelligence using visual data. However, it is noteworthy that spatial intelligence exists irrespective of sensory modality, even a blind person can perceive space through other senses \citep{Gardner1983FramesOfMind}. Regrettably, whether VLMs and LLMs can comprehend viewpoint rotation (i.e., VRU) without visual information still remains underexplored. To address this gap, we first propose a simple textual viewpoint rotation understanding dataset \textbf{VRUBench}, in which models are provided only with textual descriptions of multi-step viewpoint rotations and their associated observations, and are required to predict the observation after the final rotation (as illustrated in Figure \ref{fig:illustration}.). Results show that human can easily achieve 100\% accuracy on VRUBench while the majority of current LLMs and VLMs perform poorly. Even state-of-the-art (SOTA) models such as Qwen3-VL \citep{bai2025qwen3vltechnicalreport}, which demonstrates remarkable performance across numerous challenging tasks, achieves only $\approx60\%$ accuracy, highlighting a substantial gap between current model capability and the requirements of spatial intelligence. 

Despite the poor performance of both LLMs and VLMs, we still obtain the following valuable findings: i) \textbf{Firstly, VLMs consistently outperform LLMs} (e.g., Qwen3-VL-8B vs Qwen3-8B), indicating that training with visual data enhances models’ spatial perception ability, even when no visual input is available at inference time. ii) \textbf{Secondly, for both VLMs and LLMs, enabling models to ``think-then-answer'' achieves better performance than the no-reasoning setting}. This observation complements the conclusion of prior work \citep{DBLP:conf/cvpr/YangYGH0X25}, which reports that eliciting model's reasoning capability through chain-of-thought (CoT) \citep{DBLP:conf/nips/Wei0SBIXCLZ22} does not improve the performance on their \emph{visual}–spatial dataset VSI-Bench. In contrast, our results indicate that under a \emph{text}-only setting, reasoning-based methods can somewhat enhance model performance, suggesting an inherent difference between text-only and vision-based spatial perception. 

To uncover the intrinsic mechanisms underlying the poor VRU performance, which could provide insights for building advanced models with stronger spatial intelligence, we conduct comprehensive interpretability studies including layer-wise and head-wise analysis. Concretely, to evaluate the capabilities of LLMs and VLMs for encoding the information of viewpoint rotation, we conduct a layer-wise probing analysis on VRUBench. We find that models have a strong capability to encode the direction and angle of the viewpoint rotation at each step, while the capability of encoding absolute orientation gradually emerges in early-to-middle layers and diminishes in later layers. 

To figure out what happens in the later layers, a causal intervention method, known as \emph{path patching} \citep{DBLP:conf/iclr/WangVCSS23} is utilized to conduct head-wise interpretation. Through path patching, a small fraction of attention heads located in middle-to-late layers are identified as responsible for VRU performance (i.e., key heads), where the attention patterns are also interpreted in a human-understandable manner. Our findings reveal that after the model acquires the viewpoint information through early-to-middle layers, the \textbf{proposal head} extracts all possible candidate answers then the final answer is selected by \textbf{answer decision head}. We also find a special head, \textbf{unknown head}, which reflects a cautious behavior in answer prediction and may be induced by alignment training \citep{DBLP:journals/corr/abs-2502-13923, DBLP:journals/corr/abs-2504-10479}. In other words, in the later layers, the model gradually transitions from orientation perception to the decision-making stage, which explains aforementioned decline in orientation-probing accuracy. However, the poor performance in VRU demonstrates that these heads fail to effectively bind the perceived viewpoint orientation from the preceding layers with the corresponding observation at that orientation, resulting in hallucinated answer decisions. 

As aforementioned, since the key heads do not function effectively to choose correct answers, it motivates us to selectively fine-tune these heads to improve the VRU performance. Experimental results demonstrates that selective fine-tuning can achieve enhanced VRU performance with only 50\% of the GPU hours required for full fine-tuning, while preserving the general capabilities. 

\section{Related Work}
\paragraph{Spatial Intelligence}
Spatial intelligence exists irrespective of sensory modality, even a blind person can perceive space through other senses \citep{Gardner1983FramesOfMind}. Regrettably, researches on spatial intelligence primarily focus on \emph{visual}-spatial intelligence \citep{DBLP:conf/cvpr/YangYGH0X25, DBLP:conf/iclr/RamakrishnanWKK25, guo2025flatlandsunlockingspatialintelligence, DBLP:journals/corr/abs-2410-16162} while \emph{textual}-spatial intelligence still lacks a sufficient study. The most relevant existing studies have largely concentrated on benchmarking and enhancing LLM performance in understanding the \textit{spatial relations} within \emph{static, single-viewpoint} scenarios \citep{DBLP:conf/aaai/ShiZL22, DBLP:conf/aaai/LiH024a, DBLP:journals/tmlr/YamadaBLKY24}, leaving dynamic and viewpoint-shifting scenarios underexplored. This motivates our study of the proposed VRUBench, where spatial understanding must incorporate changes over time and viewpoint. Importantly, so far, no systematic research has analyzed the mechanisms underlying spatial intelligence.

\paragraph{Mechanistic Interpretability}
Interpreting the inner mechanisms of LLMs and VLMs has attracted increasing attention in recent years \citep{DBLP:journals/csur/MadsenRC23}. Despite treating models as black box, these works tend to uncover the intrinsic cause of models behaviors, such as hallucination \citep{gao2025hneuronsexistenceimpactorigin, DBLP:conf/iclr/Wang00TXDC25}, mathematical ability \citep{DBLP:conf/icml/ZhangWZC0SY24, DBLP:conf/emnlp/YuA24b, li2025improvingtemporalunderstandinglogic}, safety \citep{DBLP:journals/corr/abs-2508-19697, DBLP:conf/iclr/QiPL0RBM025, DBLP:conf/iclr/ZhouY0X0WLFL25} etc. However, the underlying mechanisms about how models achieve \emph{textual}-spatial intelligence in dynamic, viewpoint-shifting scenarios still remain mysterious, motivating the interpretability analysis in this work. Furthermore, a deeper understanding of model mechanisms has also shed light on model improvement, including better architectural design \cite{DBLP:conf/iclr/FuDSTRR23}, token/weight pruning \citep{jiang-etal-2025-dcp, gao2025weightsparsetransformersinterpretablecircuits} and post hoc enhancement \cite{DBLP:conf/nips/0002PVPW23, DBLP:conf/iclr/ChuangXLKGH24, DBLP:conf/acl/JuCY0DZL24}. Benefiting from advances in interpretability, some studies have attempted to perform targeted interventions on the behavior of specific attention heads based on interpretability results. For example, Inference-Time Intervention (ITI) \cite{DBLP:conf/nips/0002PVPW23} proposes that adding steering vectors to certain attention heads can improve the faithfulness of model outputs, while \citet{DBLP:conf/nips/YinYD24} propses to train the steering vectors on downstream task and add them to the hidden representations at selected heads. Our work generally follows the ``interpret-then-improve'' paradigm \citep{DBLP:conf/icml/ZhangWZC0SY24}, where models are not only being interpreted but also improved based on the interpretability findings. 

\section{Task Definition and Dataset}
\subsection{Task Definition}
\label{sec:task_def}
For text-only VRU, the prompt $P$ for LLMs and VLMs can be formalized as: 
\begin{equation}
    P = I \oplus O_0
  \oplus \overbrace{
      A_1 \oplus O_1 \oplus A_2 \oplus \cdots + A_n
    }^{n\ \text{steps}},
\end{equation}
where $\oplus$ denotes the string concatenation operation, $I$ is the task instruction, $O_0$, $A_i$ and $O_i$ describe the initial observation, direction and angle of $i$-th viewpoint rotation, and the observation after $i$-th rotation in the environment, respectively. Examples are provided in Figure~\ref{fig:illustration} and Appendix~\ref{app:samples}. 
After the final rotation $A_n$, models $\mathcal{M}$ are tasked to predict the corresponding observation: 
\begin{equation}
    \widehat{O}_n = \mathcal{M}(\,\cdot \mid P).
\end{equation}

In Sec.\ref{sec:eval}, we test 2 evaluation settings: \textbf{directly outputting the answer} and \textbf{think-then-answer}. For the former, we extract the first token generated by models as $\widehat{O}_n$. For the latter, models are asked to output the answers between \texttt{<ans></ans>} tags after chain-of-thought (CoT), and the $\widehat{O}_n$ is extracted from the tags, detailed in Appendix \ref{app:reasoning_prompt}. 

We adopt the accuracy of observation prediction as the evaluation metric:
\begin{equation}
    Acc=\frac{\sum_{i\in \mathcal{D}}\delta\left(\widehat{O}_{n}^{i},O_{n}^{i}\right)}{\left|\mathcal{D}\right|},
\end{equation}
where $\delta(x, y)$ is an indicator function that equals 1 for $x = y$ and 0 otherwise, $|\mathcal{D}|$ is the dataset size. 

\subsection{Dataset Details}
\label{sec:dataset_construction}
For dataset synthesis, we construct a simulated environment containing 100 objects that may appear in real indoor rooms. The viewpoint rotation angle $\theta$ here is restricted to 
$\theta \in \{0^\circ, 90^\circ, 180^\circ, 270^\circ, 360^\circ\}
$, which ensures that, after rotating to the nearest adjacent viewpoint, objects observed in the previous view do not simultaneously appear in the rotated view, thereby avoiding ambiguity in observation prediction. At every step, model randomly rotates its viewpoint by $\theta$ degrees to the left or right. If the model has never visited the resulting viewpoint before, we randomly sample one object from the 100 candidates and present it as the observation. Otherwise, we replay the previously observed object. After the final action, the ground-truth of observation is set as the object corresponding to the final viewpoint orientation if the viewpoint is visited before otherwise ``unknown''. 
Finally, the constructed dataset \textbf{VRUBench} consists of 19,591 instances, including 4,614 samples with 2-step rotations, 4,977 with 3-step rotations, 5,000 each with 4-step and 5-step. 

\begin{table*}[t]
\setlength{\tabcolsep}{2pt}
\centering
\begin{tabular}{llccccc}
\toprule
\multicolumn{2}{c}{\textbf{Models}}         & \textbf{2-step (\%)} & \textbf{3-step (\%)} & \textbf{4-step (\%)} & \textbf{5-step (\%)} & \textbf{Avg. (\%)} \\ 
\midrule
\multicolumn{1}{l|}{\multirow{8}{*}{LLMs}} & L2-7B-chat     & 5.44 &17.22&	26.24&25.64&18.90        \\
\multicolumn{1}{l|}{}                     & L3.1-8b-Instruct   & 35.39&37.71&38.66&34.74&36.65    \\
\multicolumn{1}{l|}{}                     & Q2.5-7B      & 58.00& 41.33&	36.18&	33.28&	41.89\\
\multicolumn{1}{l|}{}                     & Q2.5-14B      & 61.77  & 60.64 & 59.42 & 58.00 & 59.92     \\
\multicolumn{1}{l|}{}                     & Q2.5-32B      &  88.56&	74.20&	67.54&	62.28&	72.84   \\
\multicolumn{1}{l|}{}                     & Q3-4B (-thinking)      & 35.37 (43.84) &	31.75 (45.69)&	33.24 (45.66)&	34.40 (43.84)&	33.66 (44.77) \\
\multicolumn{1}{l|}{}                     & Q3-8B (-thinking)      & 41.66 (72.02)&	38.22 (52.82)&	40.66 (44.64)&	43.56 (40.42)&	41.02 (52.09) \\
\multicolumn{1}{l|}{}                     & Q3-32B (-thinking)     &  50.04 (81.32)& 51.94 (63.69)& 53.92 (49.44)& 56.18 (39.32)& 53.08 (57.99) \\
\midrule
\multicolumn{1}{l|}{\multirow{6}{*}{VLMs}} & Q2.5-VL-3B      & 52.73&	40.00&	33.42&	25.50&	37.62    \\
\multicolumn{1}{l|}{}                     & Q2.5-VL-7B      & 53.32 &49.11&48.38&44.22&48.67 \\
\multicolumn{1}{l|}{}                     & Q2.5-VL-32B      & 68.14& 61.94&	55.62& 54.30&	59.84  \\
\multicolumn{1}{l|}{}                     & Q3-VL-4B (-thinking)      &  47.83 (50.65) & 48.60 (45.11) & 49.06 (40.30) & 48.20 (36.98) & 48.44 (43.11)    \\
\multicolumn{1}{l|}{}                     & Q3-VL-8B (-thinking)      & 64.33 (85.57)&	61.28 (62.38)&	57.04 (55.48)&	54.58 (50.94)&	59.21 (63.16)  \\
\multicolumn{1}{l|}{}                     & Q3-VL-32B (-thinking)     &  77.50 (97.90)&	71.03 (96.44)&	67.54 (96.16)&	64.42 (95.82)&	69.98 (96.55) \\
\multicolumn{1}{l|}{}                     & G3-Flash (-thinking)     &  84.15 (93.15) & 77.22 (90.32) & 73.17 (85.71) & 69.73 (76.65) & 75.91 (86.32) \\ 

\midrule
                                          & Human Performance &  100      &  100      &   100     &   100 & 100     \\
\bottomrule
\end{tabular}
\caption{The accuracy of various LLMs and VLMs in $i$-step ($i=2,3,4,5$) viewpoint rotation tasks. We also compute the average accuracy (\textbf{Avg.}) to evaluate the overall performance. Due to table length constraints, we use L to denote \textbf{LLaMA}, Q to denote \textbf{Qwen}, and G to denote \textbf{Gemini}. }
\label{tab:benchmark}
\end{table*}

\section{Evaluation}
\label{sec:eval}
\subsection{Experimental Settings}
Since research on specialized spatial intelligence models is still in its early stages, this paper primarily focuses on generic LLMs and VLMs. For LLMs, we experiment with LLaMA2-7B-chat \citep{DBLP:journals/corr/abs-2307-09288}, LLaMA3.1-8B-Instruct \citep{DBLP:journals/corr/abs-2407-21783}, Qwen2.5 series \citep{DBLP:journals/corr/abs-2412-15115}, Qwen3 series \citep{DBLP:journals/corr/abs-2505-09388}. For VLMs, we test Qwen2.5-VL series \citep{DBLP:journals/corr/abs-2502-13923}, Qwen3-VL series \citep{bai2025qwen3vltechnicalreport} and the latest frontier model Gemini3-Flash \cite{gemini3flash}. Besides, for Qwen3 and Qwen3-VL, which support reasoning mode, the results with the \emph{thinking} mode enabled are reported as ``-thinking'', which allows us to assess the impact of explicit reasoning on VRUBench. Further details are provided in Appendix \ref{app:evaluation_details}. To ensure fairness and consistency, we strictly maintained uniform hyperparameter settings across all evaluated models. Specifically, we adopted a greedy decoding strategy for text generation, setting the \texttt{temperature} to 0.0 and \texttt{do\_sample} to \texttt{False}. Furthermore, the generation length and thinking budget constraints are set as: when thinking was enabled, the \texttt{thinking\_budget} was set to 1024 (see Appendix \ref{app:reasoning_prompt}), and the \texttt{max\_new\_tokens} was capped at 2048. In contrast, when thinking was disabled, the \texttt{thinking\_budget} was set to 0, and \texttt{max\_new\_tokens} was strictly limited to 5. All other hyperparameters not explicitly mentioned were kept at their default configurations.

\subsection{Results and Discussion}
\paragraph{LLMs/VLMs vs. Human} As shown in Table \ref{tab:benchmark}, human can easily achieve 100\% accuracy in VRU while both LLMs and VLMs perform poorly. Considering that the VRU task represents only one of the capabilities required for spatial intelligence, the overall unsatisfactory performance in VRU indicates that current models are still far from possessing genuine spatial intelligence. 
\paragraph{LLMs vs. VLMs} Apart from the human evaluation results, we also observe systematic differences across model types, that is, VLMs generally outperform LLMs. This advantage is observed not only across different model families (e.g., Qwen2.5-VL-7B vs. LLaMA2-7B), but also consistently between VLMs and their corresponding LLM backbones sharing the same architecture (e.g., Qwen2.5-VL-7B vs. Qwen2.5-7B). It demonstrates that although no visual information is available at inference time, VLMs trained with visual data still exhibit stronger VRU ability than LLMs that have never been exposed to visual data.  
\begin{tcolorbox}
\textbf{Takeaway I: Training with visual data can benefit the performance in textual spatial task.}
\end{tcolorbox}
Notably, a very recent study\cite{guo2026llmspixelsbenchmarkingspatial} has compare text-only vs. visual training on more complex visual datasets. They converted 17 complex subtasks within visual-spatial intelligence, e.g., navigation, embodied perception, into text descriptions and compared the performance of LLMs and VLMs. The final results align perfectly with our observations, that is, VLMs consistently outperform LLMs on these \textit{complex} text-only spatial tasks, which further demonstrates Takewaway I. 
\begin{figure*}[t]
    \centering
    \subfigure[Probing Framework]{\includegraphics[width=0.20\textwidth]{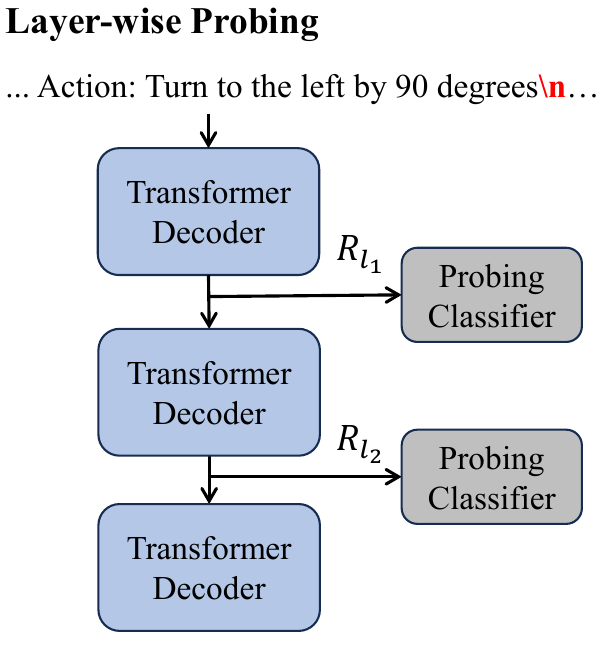}}
    \subfigure[LLaMA2-7B-chat]{\includegraphics[width=0.25\textwidth]{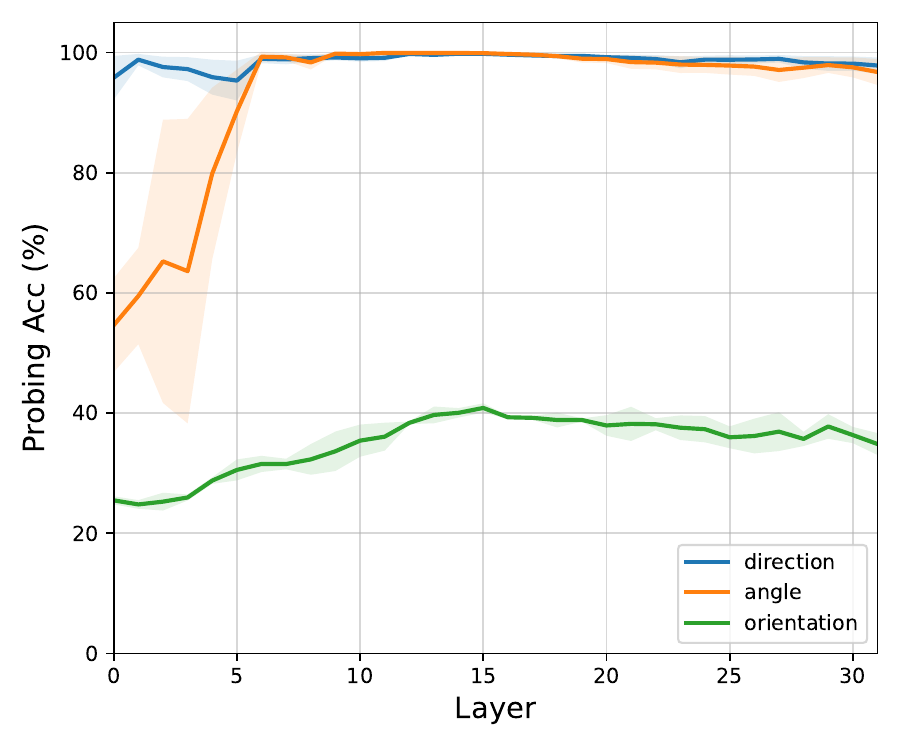}} 
    \subfigure[Qwen2.5-VL-7B]{\includegraphics[width=0.25\textwidth]{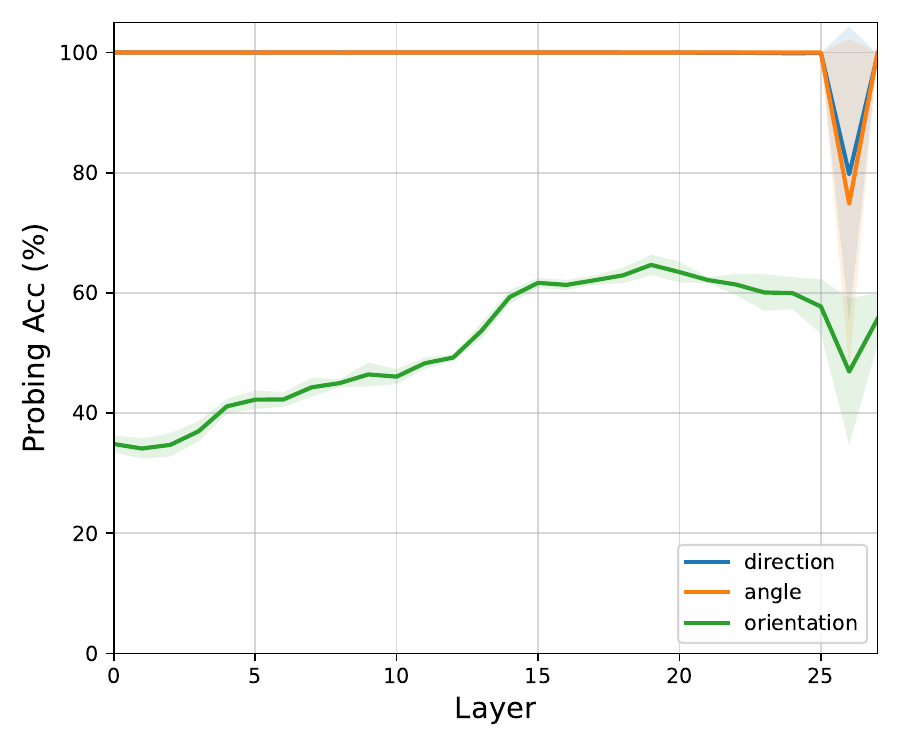}} 
    \subfigure[Path Patching]{\includegraphics[width=0.27\textwidth]{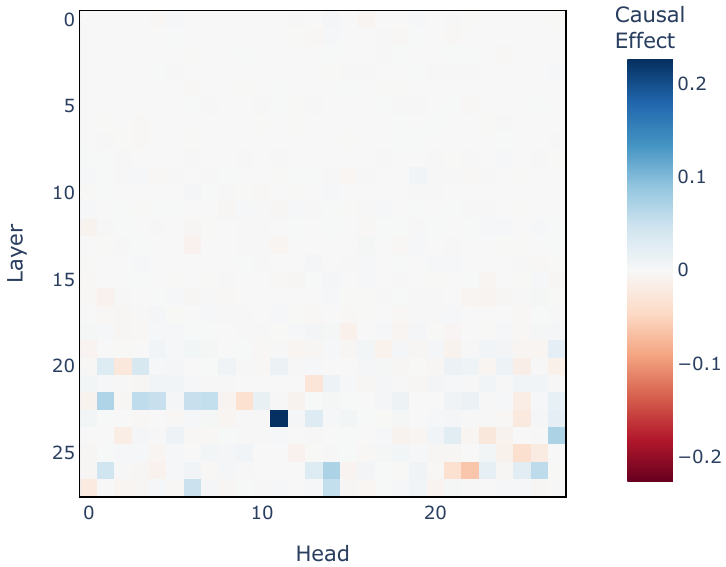}} 
    \caption{(a)-(c): Illustration of layer-wise probing and the probing results (direction/angle/orientation) on LLaMA2-7B-chat and Qwen2.5-VL-7B. (d): Path patching results of Qwen2.5-VL-7B on proposed VRUBench. 
  For each head, a darker color indicates a larger causal effect, which also reflects the importance of the head in VRU.}
\label{fig:probing_results}
\end{figure*}
\paragraph{Reasoning vs. Non-reasoning} Despite the marginal improvement in some cases (e.g., Qwen3-VL-8B-thinking outperforms Qwen3-VL-8B by only 3.95),  enabling the models to reason before outputting answers generally leads to better performance compared to directly outputting answers. Appendix \ref{app:reasoning_case} presents the model outputs when enabling the reasoning mode. It can be observed that, in this mode, models do try to reason based on the viewpoint rotation description rather than just follow templates in the data, which confirms the genuine boost to VRU task performance provided by the reasoning mechanism. Combining with the findings in \citet{DBLP:conf/cvpr/YangYGH0X25}, which shows that reasoning methods such as CoT are not effective to improve the performance on \emph{visual}-spatial intelligence, it further demonstrates the inherent difference between text-only and visual-based spatial intelligence. 
\paragraph{Effect of Scaling Law} It can be observed that the overall performance improves as the model size increases, demonstrating that the scaling law \citep{DBLP:journals/corr/abs-2001-08361, DBLP:journals/corr/abs-2203-15556} still holds in the VRU task. 

\section{How Do VLMs and LLMs Simulate Viewpoint Rotation?}
The generally poor performance of both VLMs and LLMs on the VRU task motivates us to investigate how they internally simulate viewpoint rotation, with the goal of identifying the underlying causes of their limitations. Here, we primarily focus on the mechanisms of intrinsic, implicit reasoning (i.e., under the ``directly outputting answers'' setting), which has yielded a variety of valuable findings for understanding model behaviors \citep{DBLP:conf/nips/0001LV23, DBLP:conf/icml/ZhangWZC0SY24, DBLP:journals/corr/abs-2501-14457}. Interpretability studies of the explicit reasoning process such as CoT follow another line of research, therefore fall outside the scope of this work. We leave it as a future work.

\subsection{Layer-wise Probing Analysis}
\label{sec:probing_res}
\subsubsection{Probing Method}
We first adopt the probing method to investigate whether models encode \textbf{i) the direction and angle at every rotation step}, and \textbf{ii) the absolute orientation after each rotation}. Specifically, we feed each sample in the proposed dataset into LLMs and VLMs, and extract the hidden representation associated with the last token at each action $A_i$ (the token is ``\textbackslash n'') from all layers, as illustrated in Figure \ref{fig:probing_results}(a). Subsequently, for the probing target i) and ii), we annotate each representation with different label respectively. For instance, when probing for i), the label is the direction and angle described in $A_i$. As for ii), the label is the absolute orientation after $A_i$. Data-label pairs for different probing targets can be found in Appendix \ref{app:probing_samples}. 

Finally, we obtain the representations, denoted as $R_l$ where $l \in \{1,2,...,L\}$ indexes the layer, and their corresponding direction/angle/orientation labels $Y$. Then a linear \footnote{We also explored non-linear probes such as MLPs in preliminary experiments, but observed that the conclusions drawn from probing were consistent with linear ones. Therefore, we ultimately choose to use the linear probes. Check Appendix \ref{app:probes_mlp} for further clarification.} probing classifier $\mathcal{F}_l$ is trained to map each $R_l$ to labels $Y$. By measuring the test set performance of $\mathcal{F}_l$, we can infer the extent to which the hidden layer encodes the direction/angle/orientation \citep{DBLP:conf/coling/Ju0DYRL24, ju2025probing}. 
\subsubsection{Do Models Encode Direction \& Angle?}
\label{sec:direction_angle}
As shown in Figure~\ref{fig:probing_results}(b) and (c), the probing accuracy for both direction and angle exceeds 99\% in most layers. It is not surprising since the information of rotation direction and angle is explicitly provided in $A_i$, the models could aggregate the information by self-attention mechanism \citep{DBLP:conf/nips/VaswaniSPUJGKP17} even when the texts about direction (left/right) or angle (0/90/180/270/360) are tokenized into separate tokens \citep{DBLP:conf/emnlp/YuA24b}. In summary, \textbf{LLMs and VLMs do encode the rotation direction and angle at each step}. 
\subsubsection{Do Models Encode Orientation?} 
\label{sec:orientation}
Regrettably, when probing the absolute orientation after the viewpoint rotation, as shown in Figure \ref{fig:probing_results}(b) and (c), both LLMs and VLMs struggle to achieve high accuracy across all layers. However, VLMs still exhibit overall better performance compared to LLMs, which is consistent with the better performance of VLMs in VRU. It indicates that the capability of encoding orientation is positively correlated with the model’s final performance on the VRU task. In brief, \textbf{VLMs exhibit a stronger ability to encode the information of absolute viewpoint orientation than LLMs}.

Another interesting observation is that the capability of encoding orientation within VLMs gradually emerges in early-to-middle layers (1-20) but diminishes in the later layers (21-28). Note that orientation perception is only an intermediate step toward predicting the final observation, i.e., there remains a gap between orientation and the final answer. The decrease in orientation probing accuracy therefore suggests that a pattern shift may occur in the middle-to-late layers. To verify this, we subsequently conduct a head-wise causal intervention to uncover what happened in the later layers. 

\subsection{Head-wise Causal Intervention}
\subsubsection{Path Patching}
Path patching is an interpretability technique that tests the causal role of specific internal computation paths in a model by selectively replacing activations along those paths and observing the effect on the output \citep{DBLP:journals/corr/abs-2304-05969, DBLP:conf/iclr/ZhangN24, DBLP:conf/nips/Tigges0YB24}. It typically involves the following steps: 1) running model on clean data $D_{cl}$ and corrupted data $D_{cor}$, 2) intervening targeted head's activation with corrupted one while freezing others, 3) measuring the causal effect after patching under specific metrics. 
Here, we construct the corrupted data as flipping the rotation direction at $A_n$, i.e., the last rotation step. And the metric associated with causal effect is defined as the logit difference between the answers of clean data and corrupted data:
\begin{equation}
    logit_* = \mathcal{M}(t_{cl} \mid \cdot) - \mathcal{M}(t_{cor} \mid \cdot),
\end{equation}
where $*$ denotes the clean/corrupted/patched run, $t_{cl}$ and $t_{cor}$ are the first token of the answer to clean and corrupted data, respectively. Note that when the answer does not change after flipping the rotation direction, $logit_*$ will always be $0$. To avoid this, all the clean-corrupted data pairs that share the same answers are filtered out. Finally, the causal effect is defined as: 
\begin{align}
    \phi_i&=\frac{logit_{pt}-logit_{cl}}{logit_{cor}-logit_{cl}} \\
    \Phi&=\frac{\sum_{i=1}^{|\Omega|}\phi_i}{|\Omega|},
\label{eq:causal_effect}
\end{align}
where $\phi_i$ is the causal effect for $i$-th clean-corrupted data pair, $|\Omega|$ denotes the size of such data pair after filtering, and $\Phi$ represents the averaged causal effect.
More details about path patching are provided in Appendix \ref{app:details_pp}. For simplicity, we primarily report the results of Qwen2.5-VL-7B in following discussion, while the results of other models can be found in Appendix \ref{app:more_path_patching}.

\begin{figure}
  \centering
  \includegraphics[width=0.45\textwidth]{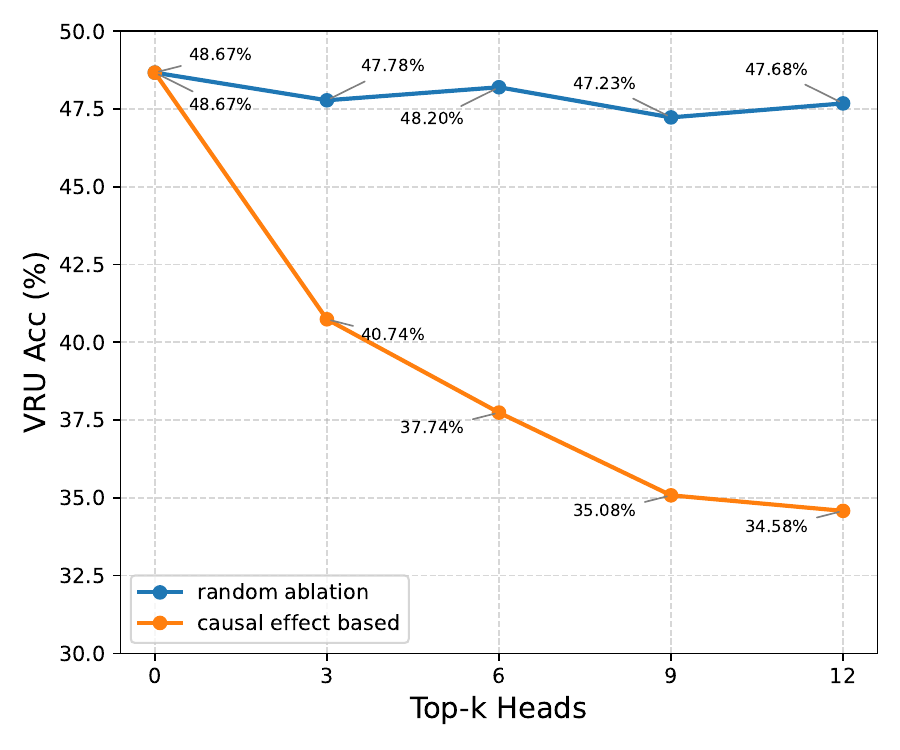}
  \caption{The VRU performance after knocking out randomly-selected $K$ heads (``random ablation''), and ablating the top-$K$ heads sorted by their causal effects.} 
  \label{fig:knock_out}
\end{figure}

\begin{figure*}[t]
    \centering
    \subfigure[Proposal Head 22.1]{\includegraphics[width=0.99\textwidth]{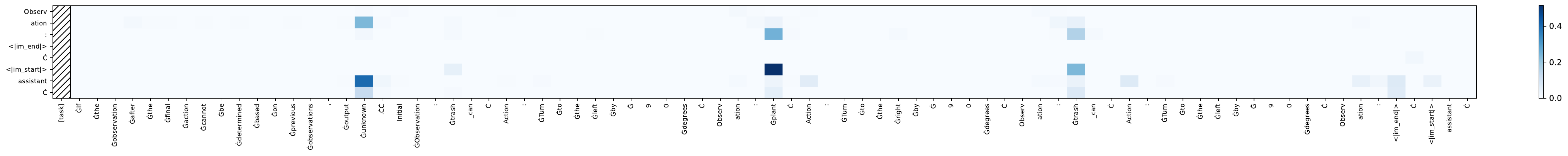}} 
    \subfigure[Answer Decision Head 26.14]{\includegraphics[width=0.99\textwidth]
    {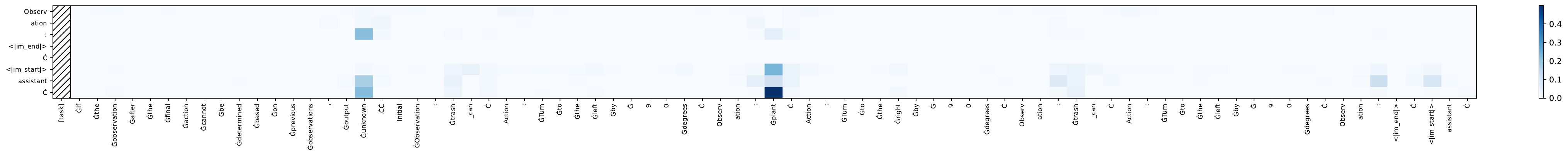}}
    \subfigure[Unknown Head 27.14]{\includegraphics[width=0.99\textwidth]{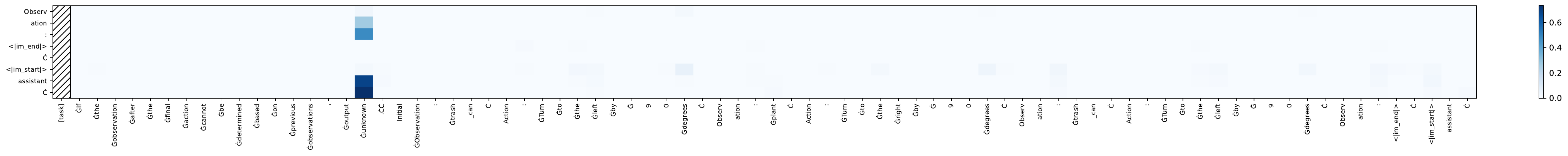}}
    \caption{The attention patterns of key heads within Qwen2.5-VL-7B, where the head index 26.14 represents the 14-th head in 26-th layer. The model's output for above example is ``plant'', the content on which the answer decision head 26.14 focuses. For ease of visualization, the task description part in the prompt is omitted and replaced with $\boxslash{}$.}
\label{fig:att-patterns}
\end{figure*}

\subsubsection{Key Heads Identification}
\label{sec:identify}
Figure \ref{fig:probing_results}(d) depicts the results of path patching (i.e., $\Phi$), where he intensity of the color reflects the strength of causal effect. We can find that only a small fraction of heads have a relatively significant effect on the answer generation. In other words, the key attention heads in VRU is particularly sparse, which is consistent with previous findings \citep{DBLP:conf/icml/ZhangWZC0SY24, DBLP:conf/naacl/YangJL25}. More importantly, \textbf{these key heads are mainly located in the middle-to-upper layers (21-28) while the heads in early layers almost exhibit zero causal effect}. Recalling that the information encoding viewpoint orientation also declines after layer 20 (Section \ref{sec:orientation}), it suggests that these key heads must play a pivotal role in this process.

\subsubsection{Validation}
\label{sec:validation}
To further validate the faithfulness of the identified key heads, we conduct additional validation experiments by measuring the performance degradation when knocking out these heads, and comparing it with randomly knocking out. Settings of the knocking out experiments are detailed in Appendix \ref{app:ablation}. Figure \ref{fig:knock_out} illustrates the results after ablating the top-k heads sorted by causal effect. It shows that, as the key heads are gradually knocked out, the performance drops significantly, while remaining stable in random ablation. 
It demonstrates that the identified heads do play an important role in VRU task, and have strong causal significance for VRU performance.
Given that the importance and faithfulness of these heads have been validated, to uncover the function of the identified heads during VRU, we subsequently conduct an attention pattern analysis to check what content the key heads attended to. 

\subsubsection{Attention Pattern Analysis}
\label{sec:att_pattern}
The sparsity of the key heads enables us to analyze their attention patterns manually. Figure \ref{fig:att-patterns} visualizes the attention patterns with specific question of the key heads identified by path patching. More results about attention pattern are supplemented in Appendix \ref{app:more_att_pattern}. Regarding the function of these key heads, it can be summarized as follows:
\paragraph{Proposal Head} We find that the head 22.1 consistently focuses on all the candidate answers in prompts, including the token ``unknown'' when the final answer cannot be determined based on the observation histories. In other words, such head acts as a proposer that extracts all the possible answers and delivers the results to subsequent components. 
\paragraph{Answer Decision Head} After receiving the candidate answers from 22.1, the answer decision head 26.14 and 23.11 (Appendix \ref{app:more_att_pattern}) precisely increases its attention to the chosen output (i.e., ``plant'') while simultaneously reducing attention to other candidates (``unknown'' and``trash\_can''), resulting in a pronounced focus on the final answer. 
\paragraph{Unknown Head} Surprisingly, even after the decision heads have precisely focused on the final answer, the subsequent heads do not directly maintain attention on the selected answer until the final output. Instead, we observe that the head 27.14 exhibits a strong attention to the token ``unknown'' just before the final layers. Moreover, we find that when the model is instructed to output an alternative token (semantically unrelated) in cases where the final observation cannot be determined, it loses such special attention. But the same pattern re-emerges when it is replaced with ``\includegraphics[width=1.0cm]{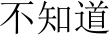}'', the Chinese translation of ``unknown'' (shown in Appendix \ref{app:unknown_head}). These observations demonstrate that the head is not simply focusing on what to respond when the answer is indeterminate. Instead, its distinctive attention to ``unknown'' and ``\includegraphics[width=1.0cm]{figs/buzhidao.pdf}'' reflects, to some extent, an inherent decision-making bias of the head. That is, a preference for cautious responses that acknowledge uncertainty rather than providing potentially incorrect definitive answers. An evidence is that after ablating the head 27.14, the overall proportion of outputting ``unknown'' significantly drops from \textbf{65.78\%} to \textbf{40.73\%}. The reason could be that after undergoing alignment training, the demand of safeguard drives the model to develop specialized attention heads that explicitly handle uncertainty by generating ``unknown'' \citep{ding-etal-2025-lvlms, DBLP:conf/naacl/ZhangDLFL0CJZ24}. 

\subsubsection{Answer Decision Mechanisms in VRU}
\label{sec:mechanisms}
In summary, the key heads in the later layers primarily function as determining the final output, indicating that the model transitions from viewpoint orientation perception to answer selection in the later layers. This finding  supports the hypothesis in Section~\ref{sec:identify} that a pattern shift does occur in the later layers, which also explains the decline of orientation-probing accuracy in those layers.

Combining with the findings in Section \ref{sec:direction_angle} and \ref{sec:orientation}, the overall internal mechanism by which models perform the VRU task can be summarized as follows:
(1) \emph{\textbf{shallow layers (sometimes even a single layer) aggregate the information of the rotation direction and angle at each step via self-attention,}} (2) \emph{\textbf{then early-to-middle layers progressively process this information to infer the absolute viewpoint orientation,}} and (3) \emph{\textbf{finally, the pattern of residual stream transforms from orientation perception to answer decision through a small subset of key attention heads at top layers.}}

Regrettably, the overall poor performance in VRU demonstrates that the key heads struggle to bind the perceived viewpoint orientation through early-to-middle layer with the observation at that orientation, resulting in hallucinated object selection conditioned on the perceived orientation. 

\begin{table*}[!t]
  \centering
  \setlength{\tabcolsep}{4pt}
  \begin{tabular}{
            @{}l @{}c @{}c  @{\hskip 10pt}
            c@{\sepsmall}c  m{0.01em} 
            c@{\sepsmall}c  m{0.01em} 
            @{\hskip 4pt}
            c@{\sepsmall}c  m{0.01em} 
            c@{\sepsmall}c 
            @{}}
      \toprule[1.25pt]
       &&&  \multicolumn{5}{c}{\textbf{Spatial Intelligence}} && \multicolumn{5}{c}{\textbf{Generic Ability}} \\ 
       
       \cmidrule(lr){4-8}
       \cmidrule(lr){10-14}
        & & & \multicolumn{2}{c}{\textbf{VRUBench}} && \multicolumn{2}{c}{\textbf{SpinBench$^*$}} &&  \multicolumn{2}{c}{\textbf{MMLU$^*$}} && \multicolumn{2}{c}{\textbf{BBH$^*$}} \\ 
        
       \cmidrule(lr){4-5}
       \cmidrule(lr){7-8}
       \cmidrule(lr){10-11}
       \cmidrule(lr){13-14}
       
        \textbf{Models}
        & \makecell[c]{\textbf{Train} \\ \textbf{Speed}} & \makecell[c]{\textbf{Tuned} \\ \textbf{Params.}} 
        &   \textbf{Acc.} & $\mathbf{\Delta}$
        &&  \textbf{Acc.} & $\mathbf{\Delta}$
        &&  \textbf{Acc.} & $\mathbf{\Delta}$
        &&  \textbf{Acc.} & $\mathbf{\Delta}$\\

      \midrule[1.25pt]
      
      Qwen2.5-VL-3B
      & -
      & -
      &  37.6 & - 
      && 46.7 & - 
      && 62.5 & - 
      && 25.5 & - \\

      \; + Full SFT 
      &  \; 10sam./sec.  
      &  3.0B  
      &  88.5  & \smallgreen{+50.9} 
      && 46.2 & \smallred{-0.5} 
      && 61.5 & \smallred{-1.0} 
      && 19.7  & \smallred{-5.8} \\ 

      \; + Selective SFT 
      &  \; 18sam./sec. 
      &  0.03B 
      &  80.1  & \smallgreen{+42.5} 
      && 47.0 & \smallgreen{+0.3} 
      && 62.9 & \smallgreen{+0.4} 
      &&  26.2 & \smallgreen{+0.7} \\ 
      
      \midrule[0.5pt]
      
      Qwen2.5-VL-7B
      & -
      & -
      &  48.7 & - 
      && 44.8 & - 
      && 60.3 & - 
      && 49.2 & - \\

      \; + Full SFT 
      &  \; 5sam./sec.  
      &  7.0B  
      &  96.3  & \smallgreen{+47.6} 
      && 47.3 & \smallgreen{+2.5} 
      && 55.6 & \smallred{-4.7} 
      && 35.8  & \smallred{-13.4} \\ 

      \; + Selective SFT 
      &  \; 11sam./sec. 
      &  0.06B  
      &  78.7  & \smallgreen{+30.0} 
      && 48.4 & \smallgreen{+3.6} 
      && 60.3 & \smallgreen{+0.0} 
      && 48.4  & \smallred{-0.8} \\

      \bottomrule[1.25pt]
  \end{tabular}
  \caption{Overall performance of selective and full fine-tuning. $^*$ represents OOD dataset. }
  \label{tab:sft}
\end{table*}

\section{Improving via Selective Fine-tuning}
\label{sec:fine-tuning}
The poor performance in VRU demonstrates that the key attention heads do not appear to be functioning effectively for choosing the correct answers. Therefore, we try to refine them by selective supervised fine-tuning (SFT). Specifically,  
selective SFT updates the parameters ($W^{i,j}_{K/Q/V/O}$) of key heads while freezing all other heads. We update the parameters of top-32 heads on constructed dataset following the same pipeline in Section \ref{sec:dataset_construction} and more details are supplemented in Appendix \ref{app:sft_detail}. 

Table \ref{tab:sft} shows the results on proposed VRUBench, along with an out-of-distribution (OOD) spatial intelligence dataset SpinBench \citep{DBLP:journals/corr/abs-2509-25390}, and generic ability benchmark MMLU \citep{DBLP:conf/iclr/HendrycksBBZMSS21} and BBH \citep{DBLP:journals/tmlr/SrivastavaRRSAF23}. With fewer tuned parameters, selective SFT achieves enhanced performance on spatial intelligence with only 50\% GPU hours compared to full SFT, while simultaneously preserving the generic abilities. Regrettably, full SFT would cause catastrophic forgetting of generic abilities despite the improvements in spatial intelligence. We provide additional discussion and analysis of catastrophic forgetting under full SFT in Appendix \ref{app:further_discussion}. Importantly, it is noteworthy that selective SFT with only \textit{textual} data even improves the performance on a \textit{visual}-spatial dataset SpinBench, which implies that: 

\begin{tcolorbox}
    \textbf{Takeaway II: Training with textual data provides transferable benefits for visual task performance.}
\end{tcolorbox}
Combining with \textbf{Takeaway I}, it demonstrates that the learning in visual and textual modalities can mutually enhance each other, which is consistent with the dual-coding theory---the idea that linguistic and visual processing are distinct yet complementary \citep{clark1991dual}. 

\section{Conclusion}
In this paper, we focus on a fundamental and critical ability in spatial intelligence: viewpoint rotation understanding (VRU). Firstly, we propose a synthized dataset to benchmark the VRU capability of both VLMs and LLMs, which yields limited performance. Through layer-wise and head-wise interpretability analysis, we not only uncover the internal mechanisms about how models performs VRU without vision, but also identify a small subset of attention heads that play a key role in this process. Beyond interpretability, the enhanced performance by selectively fine-tuning the key heads further demonstrates the potential of our findings in model improvement. Our work serve as an initial mechanistic interpretability of spatial intelligence, laying a solid foundation for future studies on more complex tasks in spatial intelligence.

\section*{Limitations}
This work conducts comprehensive interpretability studies and provides valuable insights for the spatial intelligence capability of current LLMs and VLMs. Despite these, there are still some limitations of our work, which could be summarized into the following two aspects. Firstly, LLMs and VLMs are demonstrated to be sensitive to the prompt phrasing, exploring the sensitivity in VRU task may yield additional findings. However, this falls beyond the primary scope of this paper, thus we leave it for future works. 
Secondly, due to computational resource constraints, the selective and full fine-tuning were conducted on models no larger than 7B. We hope that future research could extend it to larger-scale models. 

\section*{Ethics Statement} 
This paper explores the mechanisms underlying a textual-spatial intelligence ability and proposes to improves it with selective fine-tuning. The 100 objects used in constructing the proposed dataset, VRUBench, are manually reviewed to ensure that they do not contain any personally identifying information or malicious content. Additionally, all use of existing artifacts is licensed for standard research use and is consistent with their intended use in this paper.

The inner mechanisms uncovered in this paper provide insight for understanding model behaviors in spatial intelligence. Nonetheless, the findings in this paper may be misused to maliciously manipulate model behavior at inference time, thus we emphasize the importance of monitoring LLMs' behavior during inference. Furthermore, we do not see any other potential risks. 

\bibliography{custom}
\clearpage

\appendix
\section{Samples From VRUBench}
\label{app:samples}
\begin{tcolorbox}[colback=gray!10, colframe=black, title=\textbf{2-step rotation}]
\textbf{content}: \\ You are standing inside a room and can observe objects from your current viewpoint. You will take actions to rotate your viewpoint to the left or right, after which both your viewpoint and the objects in view changes accordingly. The actions taken are specified by ``Action:'', and the corresponding objects you see are specified by ``Observation:''. After taking the final action, predict the resulting observation. If the observation after the final action cannot be determined based on previous observations, output unknown. \\
\\
Initial Observation: avocado\\
Action: Turn to the right by 270 degrees\\
Observation: router\\
Action: Turn to the left by 90 degrees\\
Observation: \\
\textbf{ground-truth}: unknown \\
\textbf{Qwen2.5-VL-7B}: \green{unknown}
\end{tcolorbox}

\begin{tcolorbox}[colback=gray!10, colframe=black, title=\textbf{3-step rotation}]
\textbf{content}: \\ < task description > \\
Initial Observation: receiver\\
Action: Turn to the right by 180 degrees\\
Observation: clock\\
Action: Turn to the left by 180 degrees\\
Observation: receiver\\
Action: Turn to the right by 0 degrees\\
Observation: \\
\textbf{ground-truth}: receiver \\
\textbf{Qwen2.5-VL-7B}: \red{clock}
\end{tcolorbox}

\begin{tcolorbox}[colback=gray!10, colframe=black, title=\textbf{4-step rotation}]
\textbf{content}: \\ < task description > \\
Initial Observation: plate\\
Action: Turn to the left by 90 degrees\\
Observation: faucet\\
Action: Turn to the right by 180 degrees\\
Observation: wardrobe\\
Action: Turn to the left by 360 degrees\\
Observation: wardrobe
\end{tcolorbox}
\begin{tcolorbox}
Action: Turn to the left by 90 degrees\\
Observation: \\
\textbf{ground-truth}: plate \\
\textbf{Qwen2.5-VL-7B}: \red{unknown}
\end{tcolorbox}

\begin{tcolorbox}[colback=gray!10, colframe=black, title=\textbf{5-step rotation}]
\textbf{content}: \\ < task description > \\
Initial Observation: avocado\\
Action: Turn to the right by 270 degrees\\
Observation: router\\
Action: Turn to the left by 90 degrees\\
Observation: remote\\
Action: Turn to the right by 0 degrees\\
Observation: remote\\
Action: Turn to the right by 180 degrees\\
Observation: avocado\\
Action: Turn to the right by 270 degrees\\
Observation: \\
\textbf{ground-truth}: router\\
\textbf{Qwen2.5-VL-7B}: \green{router}
\end{tcolorbox}

\section{Evaluation Details}
\label{app:evaluation_details}
\subsection{System Prompt}
For the prompt template, we utilize the default system prompts of LLaMA, Qwen models. For instance, the final messages for Qwen2.5-VL-7B are as follows:
\begin{tcolorbox}
"\textbf{role}": "\textbf{system}" \\
"\textbf{content}": \\ "You are a helpful assistant"\\
"\textbf{role}": "\textbf{user}" \\
\textbf{content}: \\ < task description > \\
< Actions and Observations >
\end{tcolorbox}

\subsection{Implementation Deatails of Reasoning Mode}
\label{app:reasoning_prompt}
When the reasoning mode of Qwen3 and Qwen3-VL is enabled (denoted as ``-thinking'' in Table \ref{tab:benchmark}), the hyperparameter thinking\_budget is set to 1024, indicating that Qwen3 models generate their reasoning content of up to 1024 tokens between the \texttt{<think>} and \texttt{</think>} tokens before producing the final answer. As specified in Section \ref{sec:task_def}, in order to accurately extract the final answer generated by models, we instruct Qwen3 and Qwen3-VL to output their answers between \texttt{<ans></ans>} tags:
\begin{tcolorbox}[colback=gray!10, colframe=black, title=\textbf{Prompt for think-then-answer}]
"\textbf{role}": "\textbf{system}" \\
"\textbf{content}": \\ "You are a helpful assistant"\\
"\textbf{role}": "\textbf{user}" \\
\textbf{content}: \\ You are standing inside a room and can observe objects from your current viewpoint. You will take actions to rotate your viewpoint to the left or right, after which both your viewpoint and the objects in view changes accordingly. The actions taken are specified by ``Action:'', and the corresponding objects you see are specified by ``Observation:''. After taking the final action, predict the resulting observation. If the observation after the final action cannot be determined based on previous observations, output unknown. \textbf{\blue{Output your final answer wrapped in <ans> and </ans> tags.}} \\
\\
< Actions and Observations >
\end{tcolorbox}
We evaluate most of the models in Table~\ref{tab:benchmark} using the official Qwen API, except Qwen3-VL-4B (-thinking), for which no official API is available. When running Qwen3-VL-4B locally, we modify the prompt as follows, using the official implementation designed to elicit the reasoning capability of Qwen3 models \citep{DBLP:journals/corr/abs-2505-09388, bai2025qwen3vltechnicalreport}:
\begin{tcolorbox}[colback=gray!10, colframe=black, title=\textbf{Prompt for Qwen3-VL-4B}]
"\textbf{role}": "\textbf{system}" \\
"\textbf{content}": \\ "You are a helpful assistant"\\
"\textbf{role}": "\textbf{user}" \\
\textbf{content}: \\ \textbf{\blue{/think}} You are standing inside a room ... Output your answer between <ans></ans> tags. \\
\\
< Actions and Observations >
\end{tcolorbox}

To control the length of thinking, we reproduced the implementation related to the thinking\_budget hyperparameter, where the model is forced to terminate the thinking phase by inserting the \texttt{</think>} token to generated content once the thought content reaches a length of 1024 tokens.

\subsection{Reasoning Examples}
\label{app:reasoning_case}
Figure \ref{fig:cot_case_1}-\ref{fig:cot_case_2} present the results when thinking mode is enabled. It can be observed that the models extract information from input and try to solve the problems from a mathematical perspective, ultimately leading to the improved performance in Table \ref{tab:benchmark}. 

\subsection{Human Performance}
The human performance in Table \ref{tab:benchmark} are obtained from manual assessments conducted by twenty volunteers who are master students. We randomly sample 10,000 cases from the proposed VRUBench and invite a pool of 20 human evaluators. To ensure reliability, every single case was independently evaluated by two different annotators. That is, each annotator is responsible for 1,000 cases. We evaluated the consistency of the human responses and achieved a Krippendorff's Alpha \cite{krippendorff2011computing} of 1.0, which strongly demonstrates the inter-annotator reliability.

\section{Layer-wise Probing}

\begin{figure*}[t]
    \centering
    \subfigure[LLaMA3.1-8B]{\includegraphics[width=0.24\textwidth]{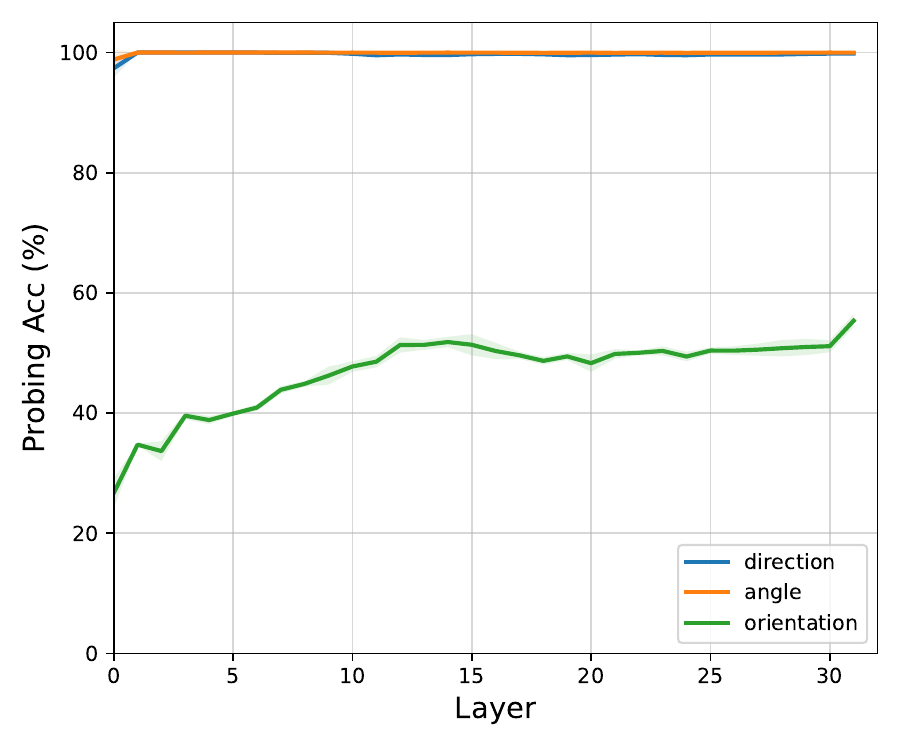}}
    \subfigure[Qwen2.5-7B]{\includegraphics[width=0.24\textwidth]{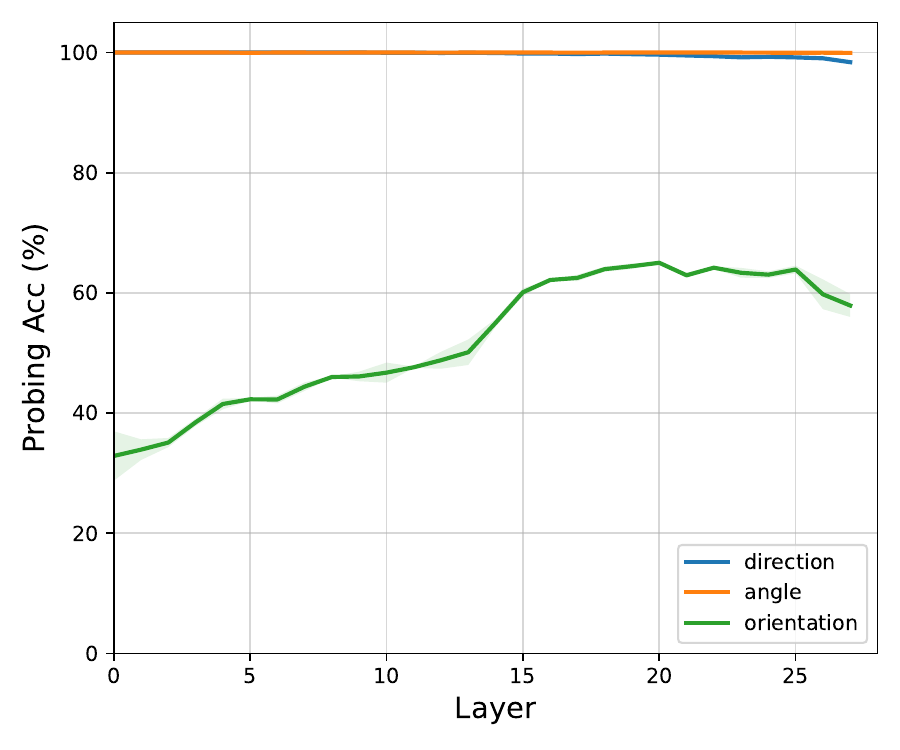}}
    \subfigure[Qwen2.5-VL-3B]{\includegraphics[width=0.24\textwidth]{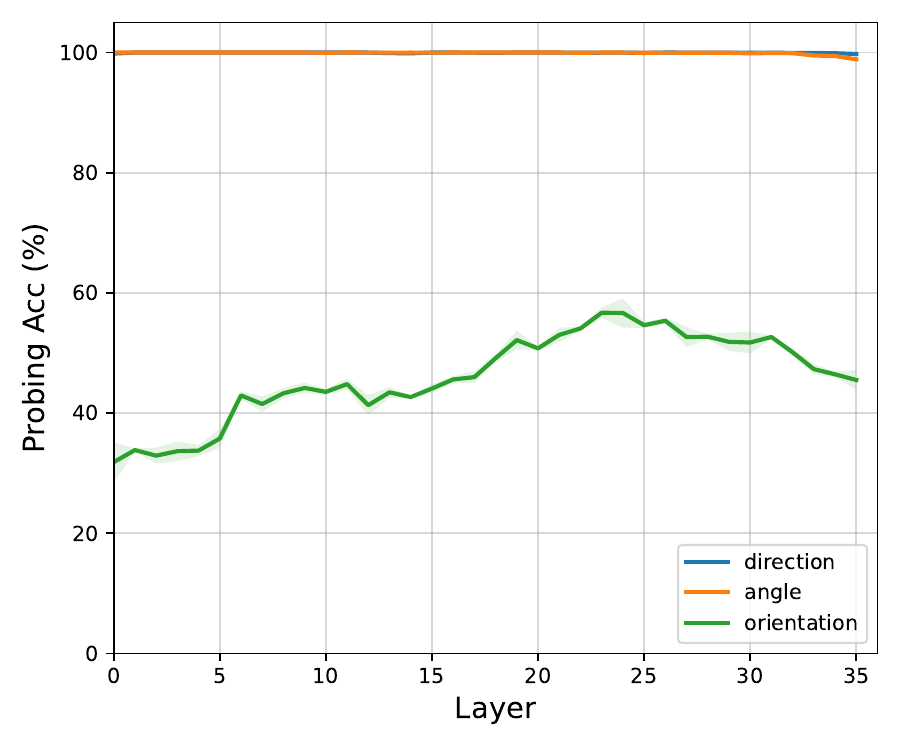}} 
    \subfigure[Qwen3-VL-4B]{\includegraphics[width=0.24\textwidth]{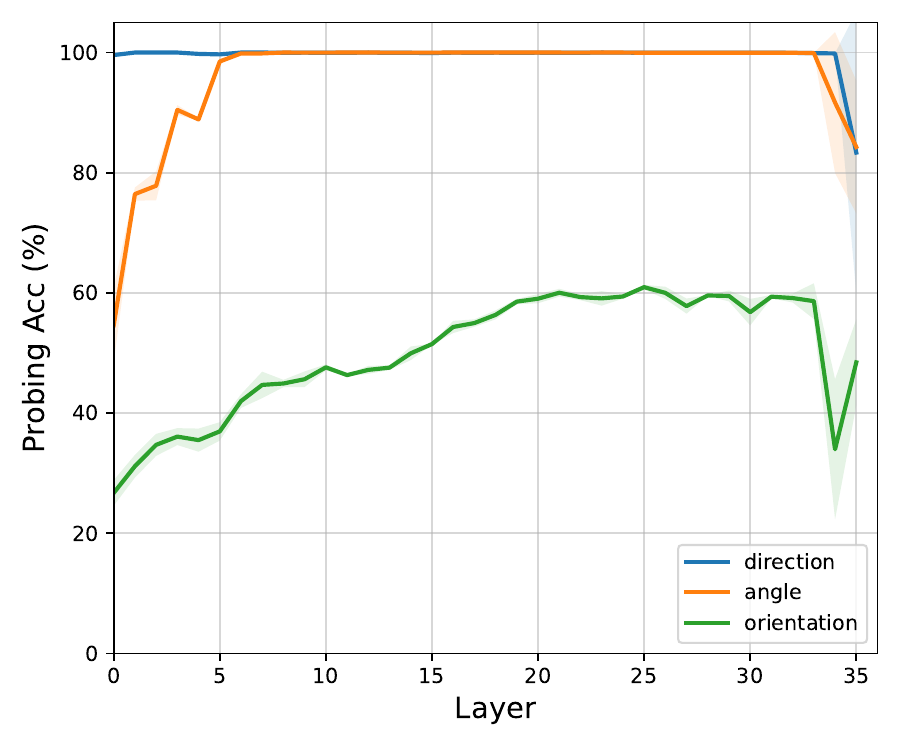}} 
     
    \caption{The probing results of other LLMs and VLMs.}
\label{fig:app_probing_results}
\end{figure*}

\subsection{Dataset for Probing}
\label{app:probing_samples}
For probing \textit{direction} (the probing label is shown after $\triangleright$):
\begin{tcolorbox}[colback=gray!10, colframe=black, title=\textbf{Labels for Probing Direction}]
\textbf{content}: \\ < task description > \\
Initial Observation: avocado\\
Action: Turn to the right by 270 degrees\\ \LineComment{\blue{right}} \\
Observation: router\\
Action: Turn to the left by 90 degrees\\ \LineComment{\blue{left}} \\
Observation: 
\end{tcolorbox}

For probing \textit{angle}:
\begin{tcolorbox}[colback=gray!10, colframe=black, title=\textbf{Labels for Probing Angle}]
\textbf{content}: \\ < task description > \\
Initial Observation: avocado\\
Action: Turn to the right by 270 degrees\\ \LineComment{\blue{$270^\circ\ $}} \\
Observation: router\\
Action: Turn to the left by 90 degrees\\ \LineComment{\blue{$90^\circ\ $}} \\
Observation: 
\end{tcolorbox}

For probing \textit{orientation}:
\begin{tcolorbox}[colback=gray!10, colframe=black, title=\textbf{Labels for Probing Orientation}]
\textbf{content}: \\ < task description > \\
Initial Observation: avocado\\
Action: Turn to the right by 270 degrees\\ \LineComment{\blue{$270^\circ\ $}} \\
Observation: router\\
Action: Turn to the left by 90 degrees\\ \LineComment{\blue{$180^\circ\ $}} \\
Observation: 
\end{tcolorbox}
When training the probes, direction prediction is formulated as a binary classification task (left/right), angle as a 5-way classification task ($0^\circ\ $/$90^\circ\ $/$180^\circ\ $/$270^\circ\ $/$360^\circ\ $), and orientation as a 4-way classification task (the four cardinal directions, indexed with $0^\circ\ $/$90^\circ\ $/$180^\circ\ $/$270^\circ\ $). We train each probe three times and report the mean and standard deviation of the results in Figure \ref{fig:probing_results} and \ref{fig:app_probing_results}.

\subsection{Probing Architecture}
\label{app:probes_mlp}
In the main experiments of probing, we utilize linear classifiers as probes to explore whether LLMs and VLMs encode direction/angle/orientation, Here, we provide further clarification about the choice of linear probes rather than non-linear ones. 

Theoretically, linear probes impose a minimal inductive bias and therefore serve as a more faithful diagnostic of what is already encoded in the representation \citep{alain2018understandingintermediatelayersusing}. In contrast, non-linear probes may introduce extra expressive capacity that allows the probe to solve the task independently, thereby confounding the interpretation of probing accuracy as evidence of representational content. 

Empirically, in our preliminary experiments, we find that although non-linear probes could improve the overall probing accuracy across various LLMs and VLMs, the conclusions drawn from the probing keep consistent with linear probes. For instance, the probing accuracy of LLaMA2-7B and Qwen2.5-VL-7B in Figure \ref{fig:probing_results} would increase simultaneously when probing with MLPs. However, the overall accuracy of Qwen2.5-VL-7B still outperforms LLaMA2-7B, and increases across layers 1-20 while declining over layers 21-28, which aligns with the findings using linear probes (Section \ref{sec:probing_res}). That is, non-linear probes do not affect the conclusion of probing, which is also consistent with findings in previous study \citep{ju2025probing}. 

Therefore, based on aforementioned theoretical and empirical analysis, we ultimately opt to use linear probes during the main experiments.

\subsection{Probing Results on Other LLMs and VLMs}
\label{app:more_probing}
The probing results of other LLMs (LLaMA3.1-8B and Qwen2.5-7B) and VLMs (Qwen2.5-VL-3B and Qwen3-VL-4B) are shown in Figure \ref{fig:app_probing_results}, which exhibit consistent observations with Figure \ref{fig:probing_results}. 

\begin{figure*}
  \centering
  \includegraphics[width=0.98\textwidth]{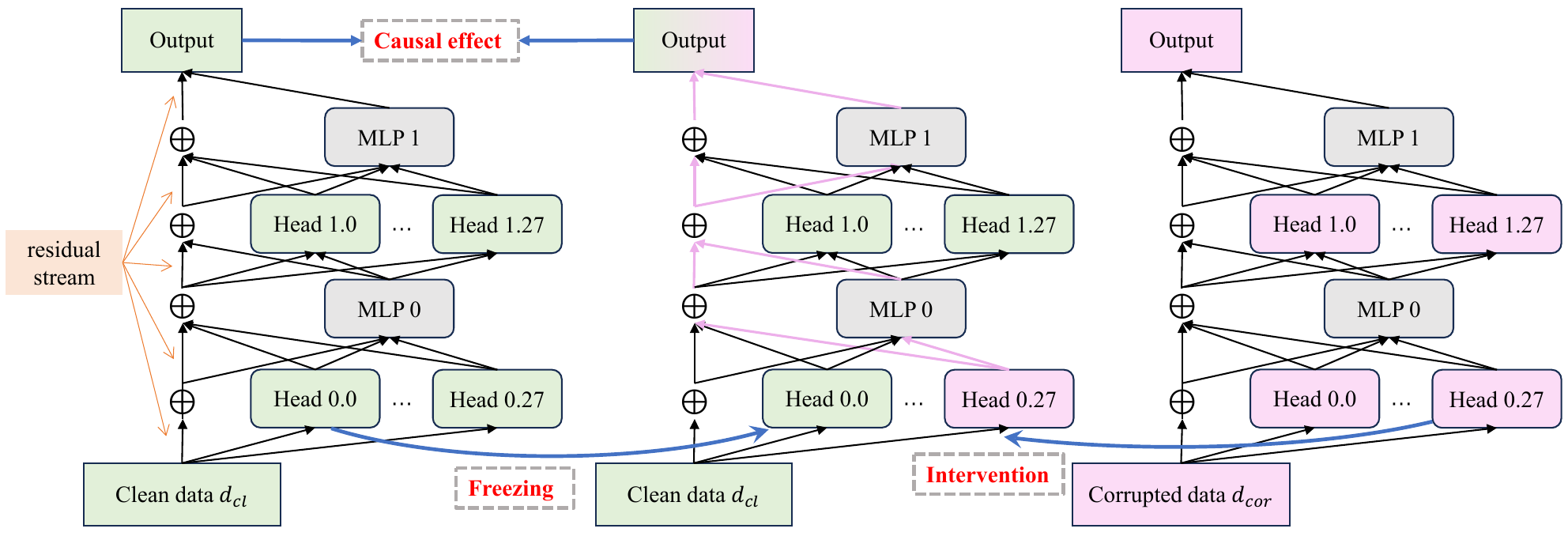}
  \caption{An illustration of the practical application of \emph{path patching}. } 
  \label{fig:illuatration_pp}
\end{figure*}

\begin{figure*}[t]
    \centering
    \subfigure[Perturbing the observation]{\includegraphics[width=0.32\textwidth]{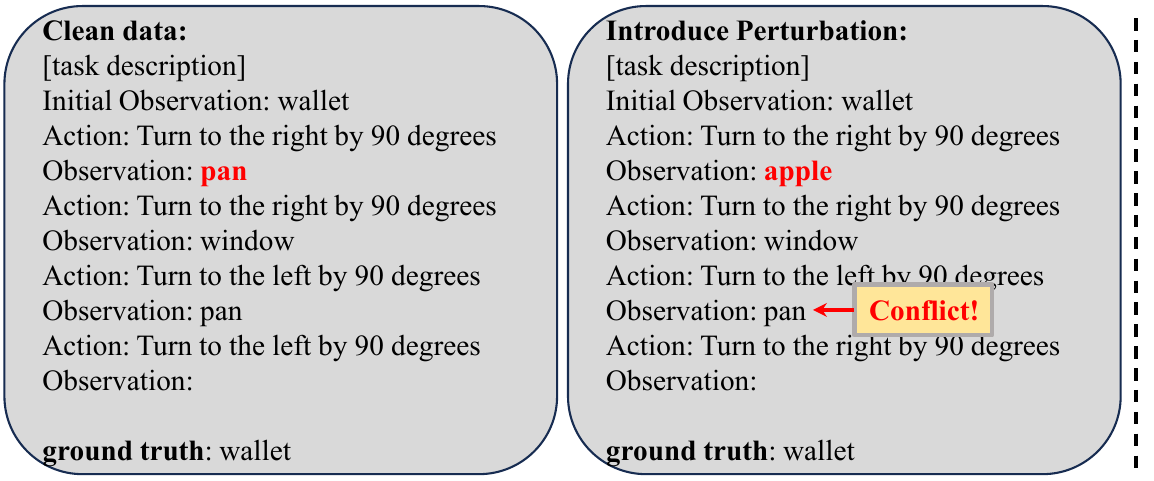}}
    \subfigure[Perturbing the direction at early step]{\includegraphics[width=0.32\textwidth]{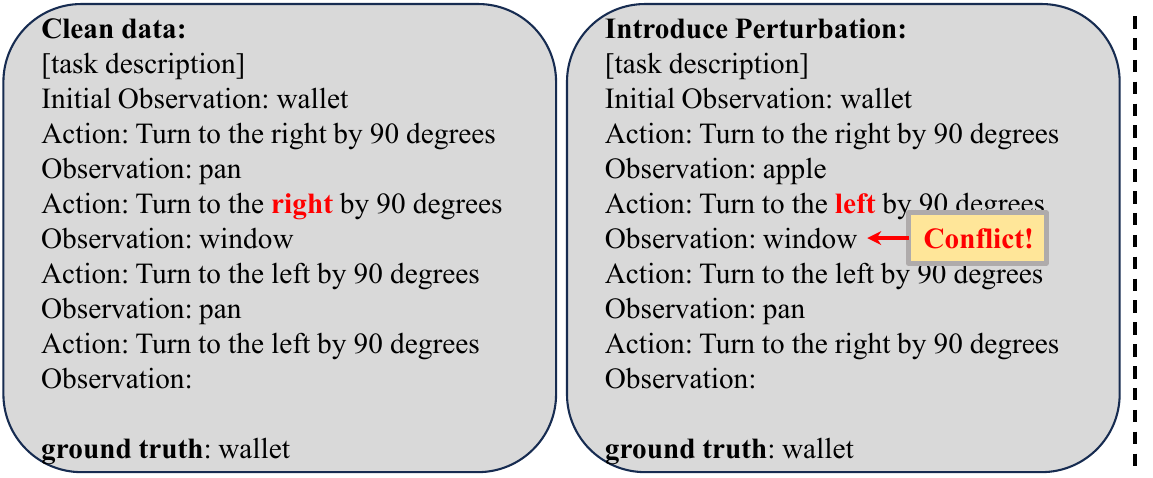}}
    \subfigure[Perturbing the direction at last step]{\includegraphics[width=0.32\textwidth]{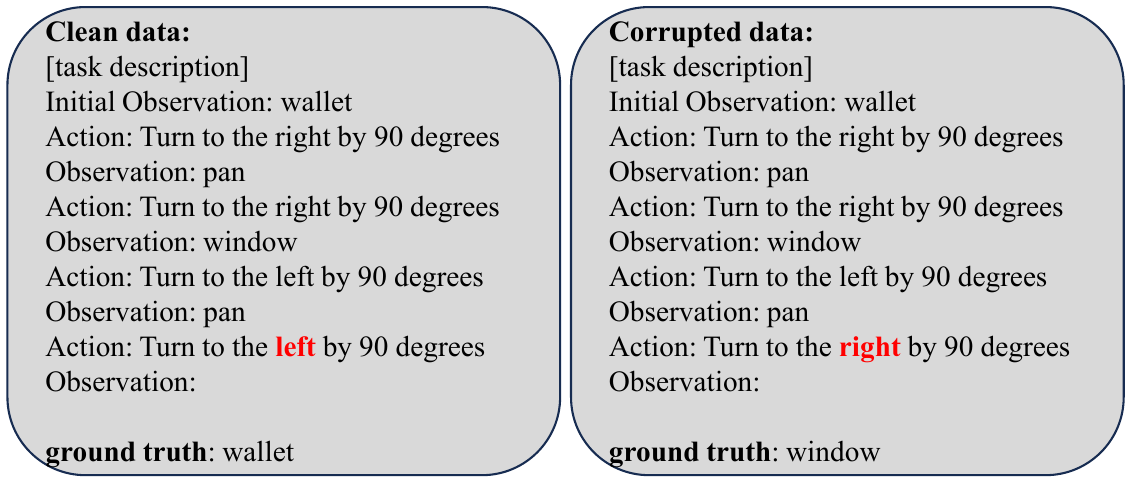}} 

    \caption{Perturbing the observation (a) or rotation direction at early step (b) could introduce logical inconsistencies in the textual description. Hence, we perturb the rotation direction at the last step to avoid such conflicts and construct the clean-corrupted data pairs (c).}
\label{fig:corr_data_construction}
\end{figure*}

\section{Details about Path Patching}
\label{app:details_pp}
Transformers \citep{DBLP:conf/nips/VaswaniSPUJGKP17} can be conceptualized such that the residual connections form the main information pathway (the residual stream, illustrated in Figure \ref{fig:illuatration_pp}), while the attention modules and MLPs act as bypass pathways that add their computations to the residual stream. Path patching is a causal intervention technique designed to identify and quantify the contribution of specific internal computational pathways to a model’s final prediction \citep{DBLP:journals/corr/abs-2304-05969, DBLP:conf/iclr/WangVCSS23}. The core idea is to treat internal components (e.g., attention heads, MLP blocks, or residual streams) as nodes in a computational graph and to test whether information transmitted along two nodes (Sender $\xrightarrow{}$ Receiver) is causally responsible for the model’s behavior.

Formally, given a clean input $d_{cl}$ and a corrupted input $d_{cor}$, path patching performs controlled interventions on selected intermediate activations while holding all other components fixed, as depicted in Figure \ref{fig:illuatration_pp}. Let $h$ denote a Sender node (e.g., an attention head) and 
$y$ denote a Receiver node (e.g., the output logits). The goal is to determine whether information flowing along the path $h \xrightarrow{} y$ is necessary for producing the observed output.

The procedure consists of three main steps:
\paragraph{Activation collection.} The model is first run on both $d_{cl}$ and $d_{cor}$, and the activations of all relevant internal components are recorded. These cached activations serve as sources for later interventions. 

\paragraph{Selective intervention.} A hard intervention is applied to the sender node 
$h$ by replacing its activation on $d_{cl}$ with the activation on $d_{cor}$, where the effect will be further propagated to the Receiver node along with a set of computational paths. Simultaneously, all other components not on the tested path are frozen to their activations under $d_{cl}$. This isolates the causal effect of the sender node by ensuring that the only information differing between the two runs flows through the designated path. 

\paragraph{Effect measurement.} The modified activations are propagated forward through the network to produce new output logits. The causal importance of the path $h \xrightarrow{} y$ is then quantified by the causal effect metrics (e.g., Eq.\ref{eq:causal_effect}). If intervening on $h$ induces a substantial change in the output while all other components remain fixed, this provides evidence that the path $h \xrightarrow{} y$ is causally important for the model’s computation on the task.

\begin{figure*}[t]
    \centering
    \subfigure[LLaMA2-7B-chat]{\includegraphics[width=0.37\textwidth]{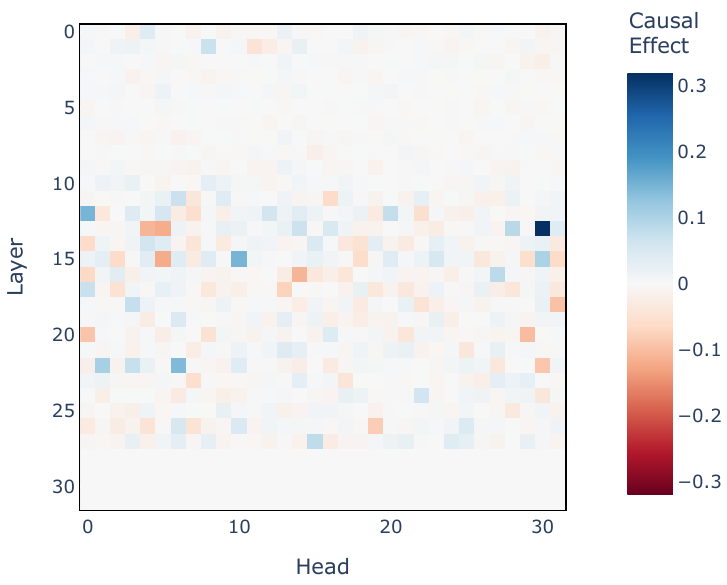}}
    \subfigure[Qwen2.5-7B]{\includegraphics[width=0.37\textwidth]{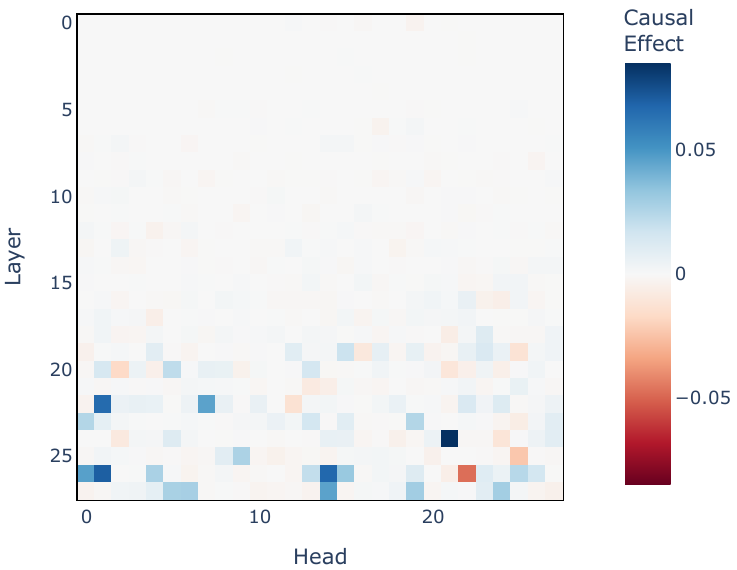}}
    \subfigure[Qwen2.5-VL-3B]{\includegraphics[width=0.24\textwidth]{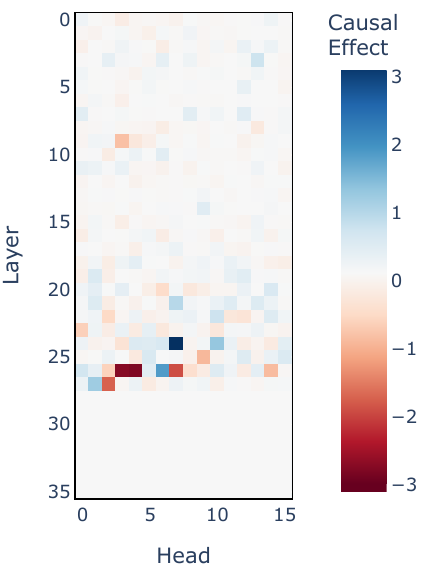}} 
     
    \caption{The results of path patching on other models.}
\label{fig:more_path_patching}
\end{figure*}
\section{More Results of Path Patching}
\label{app:more_path_patching} We further report the results of path patching on LLaMA2-7B-chat, Qwen2.5-7B and Qwen2.5-VL-3B in Figure \ref{fig:more_path_patching}. It can be observed that these models with different model families and sizes exhibit similar path patching results presented in Section \ref{sec:identify}, where key heads are distributed sparsely and in the middle-to-upper layers. 

\begin{figure*}[t]
    \centering
    \includegraphics[width=0.99\textwidth]{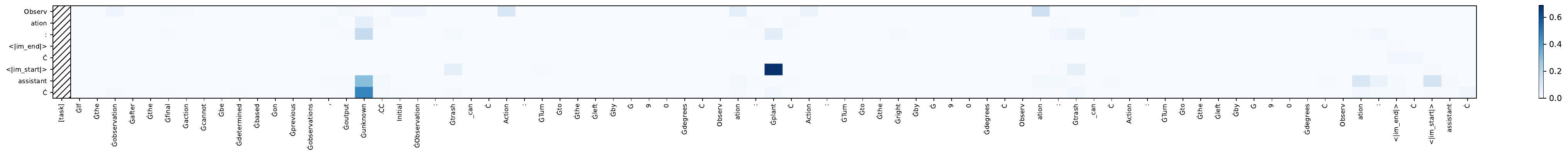} 
    \caption{Attention pattern of answer decision head 23.11, which exhibits similar attention to another answer decision head 26.14.}
\label{fig:att-patterns-23.11}
\end{figure*}

For the construction of corrupted data, path patching generally aims to apply minimal perturbation to the clean input, and ideally, such perturbations should preserve the token length to ensure comparability of internal activations. In our dataset, two elements are theoretically perturbable: the rotation (angle and direction) and the object being observed. However, rotation angles such as 0, 90 and 180° may be tokenized into sequences of different lengths, violating the token-length consistency requirement. Perturbing the observed object at any step risks introducing semantic contradictions (Figure \ref{fig:corr_data_construction}(a)), and altering the rotation direction in earlier steps can similarly result in inconsistencies across the sequence (Figure \ref{fig:corr_data_construction}(b)). To mitigate these issues, we choose to perturb the rotation direction in the final step when constructing corrupted inputs for path patching (Figure \ref{fig:corr_data_construction}(c)).

To identify critical components, we iterate over all attention heads as Sender nodes and set the output logits as the Receiver node. The causal effect is computed as the changes in the output logit of ground-truth token (Eq.\ref{eq:causal_effect}). Paths that result in a large causal effect are interpreted as essential contributors to the model’s decision, and the corresponding Sender nodes, i.e. attention heads, are considered as key heads in VRU. 

\begin{figure*}[t]
    \centering
    \subfigure[Replace ``unknown'' with ``sad'']{\includegraphics[width=0.99\textwidth]{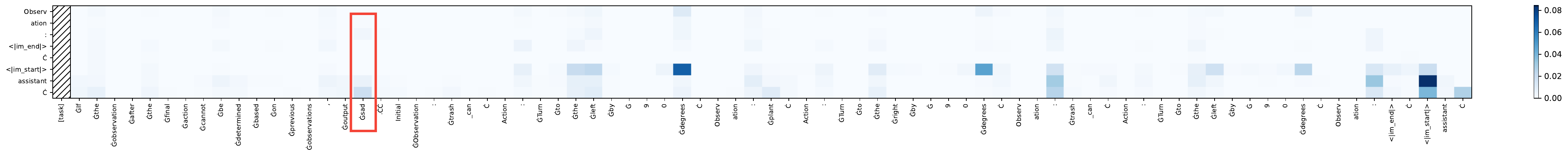}} 
    \subfigure[Replace ``unknown'' with ``cannot'']{\includegraphics[width=0.99\textwidth]
    {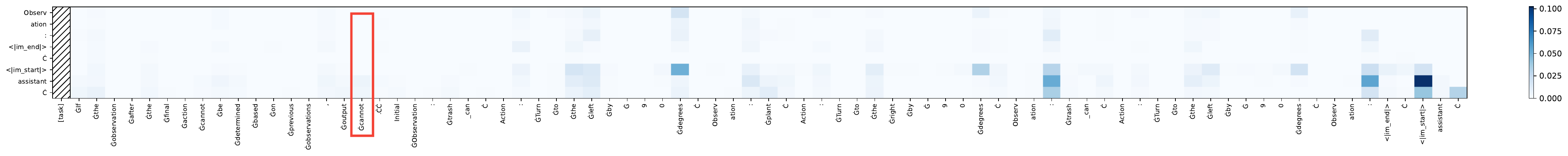}}
    \subfigure[Replace ``unknown'' with Chinese translation of ``unknown'']{\includegraphics[width=0.99\textwidth]{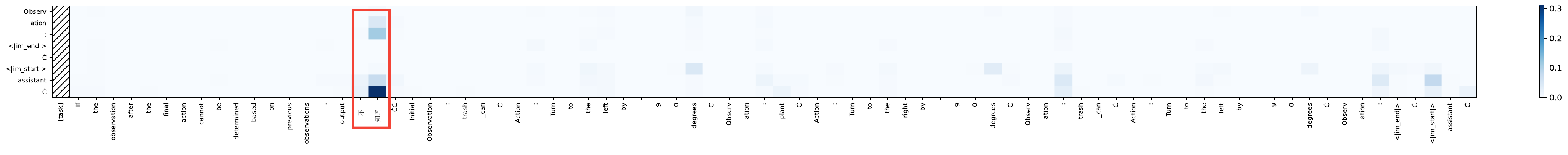}}

    \caption{Attention pattern cases of unknown head when the words ``unknown'' is replaced with other alternative tokens. The head no longer exhibits strong attention to semantically unrelated replacements such as “sad” or “cannot.” However, when “unknown” is replaced with its Chinese translation (``\includegraphics[width=1.0cm]{figs/buzhidao.pdf}''), which has the same semantic, the head again shows a similarly high level of attention to ``\includegraphics[width=1.0cm]{figs/buzhidao.pdf}''.}
\label{fig:unknown_head_other_token}
\end{figure*}

\section{Ablation Experiments}
\label{app:ablation}
In Section \ref{sec:validation} and \ref{sec:att_pattern}, we ablate the top-k key heads and the Unknown Head to validate the faithfulness of identified key heads and the function of Unknown Head, respectively. In practice, we utilize the \textbf{Undifferentiated Attention} method \citep{DBLP:conf/iclr/ZhouY0X0WLFL25} for ablation, where the parameter matrices $W_K, W_Q, W_V, W_O$ are multiplied by a very small coefficient $\epsilon$ to achieve ablation. Specifically, $W^i_K$, $W^i_Q$, $W^i_V$, and $W^i_O$  are split into $H$ blocks [$W^{i,1}_*, W^{i,2}_*, ..., W^{i,H}_*$], with each submatrix corresponding to an individual head. Here, $H$ is the number of attention heads at each layer, $W^{i,j}_K/W^{i,j}_Q/W^{i,j}_V \in \mathbb{R}^{d\times\frac{d}{H}}$, $W^{i,j}_O \in \mathbb{R}^{\frac{d}{H}\times d}$, where $d$ is the dimension of hidden representations. For ablation, the parameter matrix $W^{i,j}_K/W^{i,j}_Q/W^{i,j}_V/W^{i,j}_O$ is multiplied by a very small coefficient $\epsilon$ to diminish the contribution of a specific head in the forward computation (shown in Algorithm \ref{alg:ablation}). 

\begin{algorithm}[htbp]
\caption{Attention Head Ablation}\label{alg:ablation}
\begin{algorithmic}[1]
\State \textbf{Require}: Model $\mathcal{M}$, input $X$, index of heads to be ablated $\mathbf{\Psi}$, $W_{\theta}=W_{Q/K/V/O}$
\State \textbf{Output}: model output $\mathcal{C}$
\State \textbf{for} $(i,j) \in \mathbf{\Psi}$ \textbf{do}
\State \qquad $W_{\theta}^{i,j} \xleftarrow{}  \epsilon \, W_{\theta}^{i,j}$
\State \textbf{end for} \LineComment{Undifferentiated Attention}
\State $\mathcal{C} \leftarrow \mathcal{M}$.forward($X$)
\end{algorithmic}
\end{algorithm}

\section{More Attention Pattern Cases}
\label{app:more_att_pattern}
\begin{figure*}[t]
    \centering
    \subfigure[Proposal Head 22.1]{\includegraphics[width=0.99\textwidth]{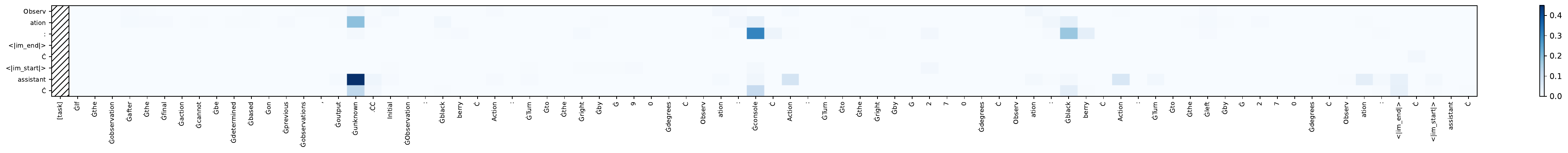}} 
    \subfigure[Answer Decision Head 26.14]{\includegraphics[width=0.99\textwidth]
    {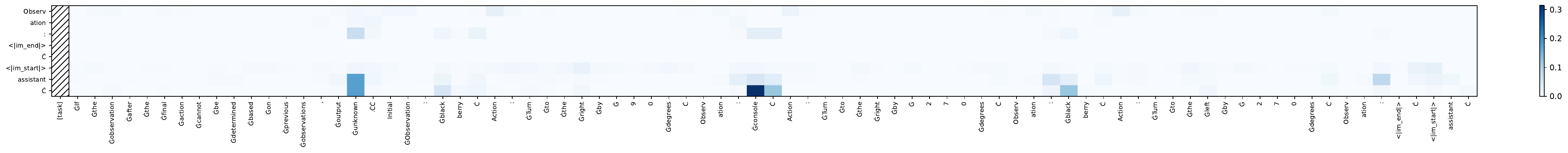}}
    \subfigure[Unknown Head 27.14]{\includegraphics[width=0.99\textwidth]{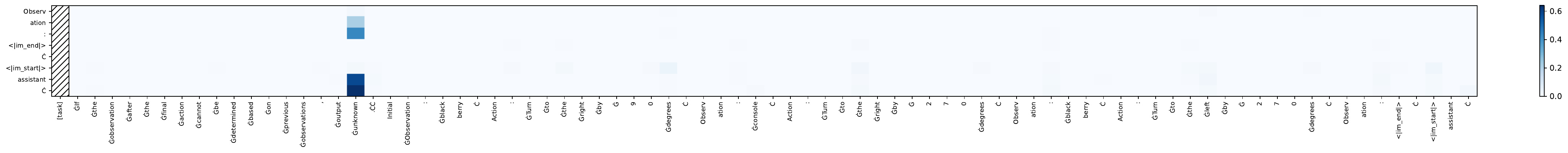}}
    \caption{\textbf{Example 1} (attention pattern). Model output: \textit{console}.}
\label{fig:att-patterns-1}
\end{figure*}

\begin{figure*}[t]
    \centering
    \subfigure[Proposal Head 22.1]{\includegraphics[width=0.99\textwidth]{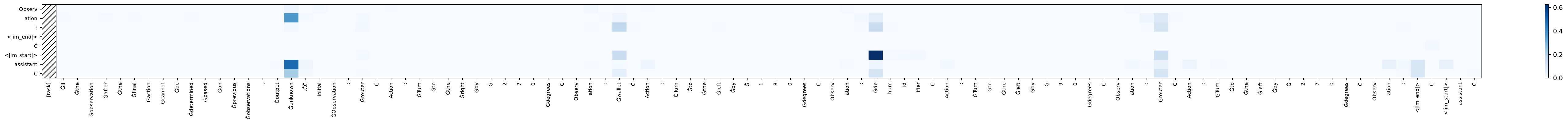}} 
    \subfigure[Answer Decision Head 26.14]{\includegraphics[width=0.99\textwidth]
    {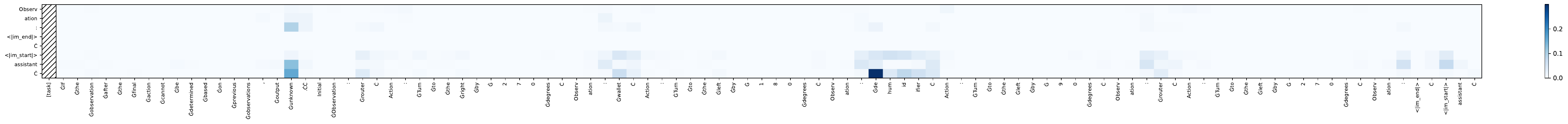}}
    \subfigure[Unknown Head 27.14]{\includegraphics[width=0.99\textwidth]{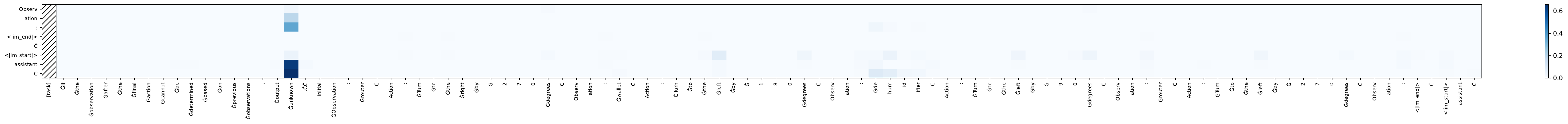}}
    \caption{\textbf{Example 2} (attention pattern). Model output: \textit{dehumidifier}.}
\label{fig:att-patterns-2}
\end{figure*}

\begin{figure*}[t]
    \centering
    \subfigure[Proposal Head 22.1]{\includegraphics[width=0.99\textwidth]{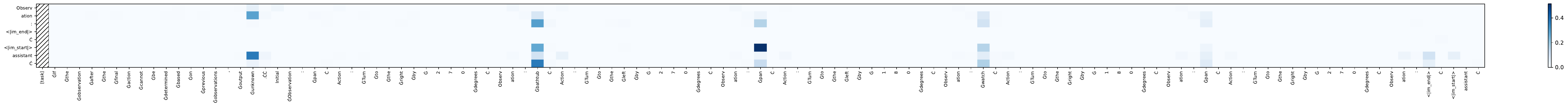}} 
    \subfigure[Answer Decision Head 26.14]{\includegraphics[width=0.99\textwidth]
    {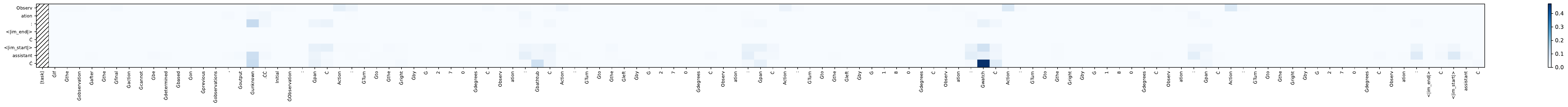}}
    \subfigure[Unknown Head 27.14]{\includegraphics[width=0.99\textwidth]{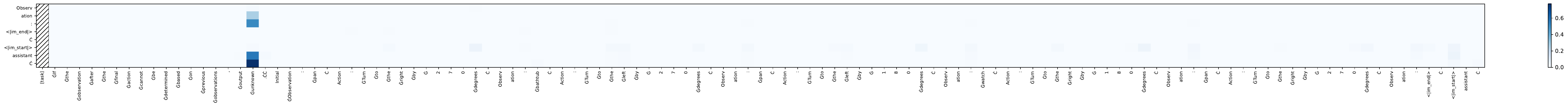}}
    \caption{\textbf{Example 3} (attention pattern). Model output: \textit{bathtub}.}
\label{fig:att-patterns-3}
\end{figure*}

The attention pattern of answer decision head 23.11 is supplemented in Figure \ref{fig:att-patterns-23.11}. And more attention pattern cases are presented in Figure \ref{fig:att-patterns-1}-\ref{fig:att-patterns-3}, where these heads exhibit consistent behaviors with those observed in Figure \ref{fig:att-patterns}. 

\section{Unknown Head with Alternative Token}
\label{app:unknown_head}
As discussed in Section \ref{sec:att_pattern}, we instruct the models to output other words instead of ``unknown'' when the final observation cannot be determined by previous observations, in order to comprehensively understand the mechanisms of the Unknown Head 27.14. As shown in Figure \ref{fig:unknown_head_other_token}, when the word ``unknown'' is replaced with semantically unrelated tokens, the head 27.14 does not keep the same particular attention on the alternative tokens, but keeps a high attention to ``\includegraphics[width=1.0cm]{figs/buzhidao.pdf}'', reasonably demonstrating the function of the unknown head as discussed in Section \ref{sec:att_pattern}. That is, the head indeed encodes a cautious response strategy, namely, acknowledging that the answer is unknown rather than simply attending to the token when the answer cannot be determined based on observation history. 

\section{Details of Selective Fine-tuning}
\subsection{Experimental Details}
\label{app:sft_detail}
\paragraph{Algorithm} Firstly, we split the parameter matrices $W^i_{K/Q/V/O}$ into $H$ blocks, which is similar to head ablation in Appendix \ref{app:ablation}. To achieve selective fine-tuning, only the parameter matrices $W^{i,j}_{K/Q/V/O}$ associated with key attention heads $i.j$ is set to tunable, while while freezing all other parameters. Following \citet{DBLP:conf/icml/Yu0Y0L24, DBLP:conf/icml/ZhangWZC0SY24}, we rescale the gradients by a factor of $\frac{H}{h}$, where $H$ denotes the total number of heads in each layer and $h$ denotes the number of heads updated in that layer. 
\begin{table}[t]
\setlength\tabcolsep{8pt}
\centering
\setlength{\tabcolsep}{1.5mm}{
\begin{tabular}{c|c|c|c|c}
\toprule
\textbf{Methods} & \textbf{Viewpoint} & \textbf{OR}&\textbf{FR} & \textbf{OI}
\\ 
\midrule 
Full SFT & 47.3 & 51.06 & 31.91 & 86.49 \\ 
\midrule
\makecell[c]{Selective \\ SFT} & 48.4 & 53.19 & 43.62 & 91.43 \\ 
\midrule
$\Delta$ & 1.10 & 2.13 & 11.70 & 5.41\\
\bottomrule
\end{tabular} }
\caption{The results of Qwen2.5-VL-7B on various sub-tasks of SpinBench, which exhibits consistent improvements. OR, FR, OI denote Object Rotation, Face Rotation, and Object Identity respectively.} 
\label{tab:more_res_spinbench}
\end{table}

\begin{figure*}[t]
    \centering
    \subfigure[Qwen2.5-VL-7B]{\includegraphics[width=0.48\textwidth]{figs/path_patching.pdf}}
    \subfigure[Qwen2.5-VL-7B after full SFT]{\includegraphics[width=0.48\textwidth]{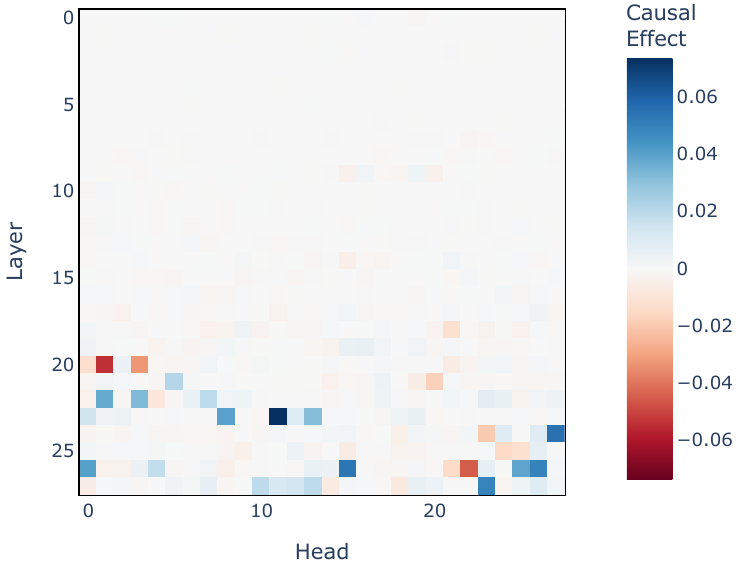}}
    \caption{The results of path patching before and after full fine-tuning.}
\label{fig:path_pathcing_after_sft}
\end{figure*}

\paragraph{Dataset} As specified in Section \ref{sec:fine-tuning}, the construction of training set for selective and full SFT follows the pipeline in Section \ref{sec:dataset_construction}, resulting in a training set comprising 19,641 samples. It consists of 4,961 2-step samples, 4,954 3-step samples, 4,996 4-step samples and 5,000 5-step samples, which is distinct from the test set VRUBench. As for another spatial intelligence benchmark SpinBench, we extract the tasks related to viewpoint as the evaluation set, which matches the research scope, i.e., viewpoint rotation understanding, in this work. Additionally, Table \ref{tab:more_res_spinbench} shows the results on extra sub-tasks of SpinBench. The selective SFT achieves consistent improvements not only on the viewpoint subset but also across other visual-spatial subtasks, which demonstrates the robustness of OOD gains and further verifies Takeaway II.
\paragraph{Hyperparameters} During selective and full SFT, we adopt
Adam \citep{DBLP:journals/corr/KingmaB14} as the optimizer with an learning rate of $2 \times 10^{-5}$. Models are trained with a batch size of $32$, warm up ratio of $0.02$ and weight decay of $0.1$ for one epoch. 

\subsection{Further Discussion about Table \ref{tab:sft}}
\label{app:further_discussion}
It is noteworthy that under selective SFT, although both Qwen2.5-VL-3B and Qwen2.5-VL-7B are trained with top-32 heads, Qwen2.5-VL-3B with smaller size surprisingly outperforms the 7B model after selective SFT. The reason could be that the 7B model consists of $28\times28=784$ heads while 3B model only contains $36\times16=576$ heads, thus 32 fine-tuned attention heads account for a lower percentage of the total heads in the 7B model. Therefore, the results do not contradict the scaling law \citep{DBLP:journals/corr/abs-2001-08361, DBLP:journals/corr/abs-2203-15556}. Another evidence is that the 7B model continues to outperform the 3B model under full fine-tuning. 

Furthermore, we also take efforts in analyzing the catastrophic forgetting of generic ability after full fine-tuning in Table \ref{tab:sft}. As shown in Figure \ref{fig:path_pathcing_after_sft}, we compare the causal effect (i.e., the importance in VRU) of each head before and after full fine-tuning. It can be observed that more heads emerge as key heads for VRU after full SFT, and some heads with relatively low importance also become prominent after full SFT. Prior studies have shown that a single attention head can simultaneously contribute to both general capabilities and task-specific functions (e.g., VRU) \citep{DBLP:conf/icml/WeiHHXQXMW024}. Therefore, the importance changes observed in Figure \ref{fig:path_pathcing_after_sft} indicate that the model transforms general-purpose heads into ones specialized for VRU, thereby resulting the extremely high performance in VRU. On the other hand, it also explains the degradation in general capabilities after full SFT.

\section{Computational Budget}
All the experiments mentioned in this paper, including the probing analysis, path patching, selective fine-tuning, are conducted on 2 NVIDIA A800 80GB GPUs. 

\begin{figure*}[h]
    \centering
    \begin{tcolorbox}[colback=gray!10, colframe=black, title=\textbf{1st Reasoning Example}]
\textbf{Input:}  \\  
Initial Observation: window\\
Action: Turn to the right by 180 degrees\\
Observation: mirror\\
Action: Turn to the left by 180 degrees\\
Observation: window\\
Action: Turn to the right by 180 degrees\\
Observation:\\
(GT: mirror) 
\tcbline
\textbf{Qwen3-8B: (pred: mirror) } \checkmark \\
Okay, let's try to figure out what the next observation should be...\\
after two turns (right and left), you're back to the original position. So if you turn right again, maybe it's the same as the first turn...\\
the previous time when you turned right from window, you saw mirror. So this seems like the same scenario. Therefore, the observation should be \green{mirror} again.
\tcbline
\textbf{Qwen3-VL-8B: (pred: mirror)} \checkmark \\
Okay, let's try to figure this out step by step.  Let me read through the initial problem again. \\
...(restate the question)...\\
Let's model the direction as angles...So a right turn of 180 degrees would be rotating 180 degrees clockwise. Similarly, left is counterclockwise.\\
\\
So, starting direction: 0 degrees (east).\\
Action 1: Turn right 180 degrees → 0 + 180 = 180 degrees (west). Observation: mirror.\\
Action 2: Turn left 180 degrees. Left turn is counterclockwise, so subtract 180. 180 - 180 = 0 degrees (east). Observation: window.\\
Action 3: Turn right 180 degrees. 0 + 180 = 180 degrees (west). Observation: \green{mirror}.\\
\\
Yes, same!
    \end{tcolorbox}
    \caption{1st reasoning example.}
    \label{fig:cot_case_1}
\end{figure*}

\begin{figure*}[h]
    \centering
    \begin{tcolorbox}[colback=gray!10, colframe=black, title=\textbf{2nd Reasoning Example}]
\textbf{Input:}  \\  
Initial Observation: cabinet\\
Action: Turn to the right by 90 degrees\\
Observation: window\\
Action: Turn to the right by 0 degrees\\
Observation: window\\
Action: Turn to the right by 90 degrees\\
Observation:\\
(GT: unknown)
\tcbline
\textbf{Qwen3-8B: (pred: unknown)} \checkmark \\
Okay, let's try to figure out what the next observation should be ...\\
Let me think: if we imagine the room as having four sides, each 90-degree turn would cycle through four directions. But the problem is that we only have two observations so far. Maybe the next observation is a different object. However, without more information, it's hard to tell. But wait, the user says that if the observation cannot be determined, output \green{unknown}.
\tcbline
\textbf{Qwen3-VL-8B: (pred: heater)} $\times$ \\
Okay, let's see. So the problem is about ...\\
So if they were facing the window (say, east), turning right 90 degrees would point them south. What's south of the window? Well, depending on the room's layout. But wait ...\\
So final observation may be \red{heater}.
    \end{tcolorbox}
    \caption{2nd reasoning example.}
    \label{fig:cot_case_2}
\end{figure*}

\end{document}